\documentclass[conference]{IEEEtran}
\usepackage{cite}
\usepackage{amsmath,amssymb,amsfonts}
\usepackage{algorithmic}
\usepackage{graphicx}
\usepackage{textcomp}
\usepackage{xcolor}
\usepackage{booktabs}
\usepackage{multirow}
\usepackage{url}
\usepackage{subcaption}
\usepackage{float}
\DeclareUnicodeCharacter{202F}{\,}
\graphicspath{{results/}}
\def\BibTeX{{\rm B\kern-.05em{\sc i\kern-.025em b}\kern-.08em
    T\kern-.1667em\lower.7ex\hbox{E}\kern-.125emX}}
\begin{document}

\title{Specificity Bias and the Wikipedia Paradox: \\ \Large An xAI Analysis of AI Text Detectors under Domain Shift}

\author{
    \IEEEauthorblockN{
        Nico Morloher\textsuperscript{1} \qquad
        Johannes Reschke\textsuperscript{1} \qquad
        Mohamed Mady\textsuperscript{1, 2}
    }
    
    \IEEEauthorblockA{
        \textsuperscript{1}\textit{Faculty of Electrical Engineering and Information Technology, OTH Regensburg, Germany} \\
        \textsuperscript{2}\textit{Technical University of Munich, Munich, Germany}
    }
    \vspace{0.1cm}
    
    \IEEEauthorblockA{
        nico.morloher@st.oth-regensburg.de \qquad
        johannes.reschke@oth-regensburg.de \qquad
        mohamed.mady@tum.de
    }
}

\maketitle

\begin{abstract}
The widespread adoption of Large Language Models (LLMs) demands reliable detection of human-authored versus synthetic text. However, current Transformer-based detectors often experience severe performance degradation during domain shifts and operate as opaque classifiers. This paper investigates the robustness and interpretability of fine-tuned Transformer models (DistilBERT, RoBERTa) by evaluating them against a custom out-of-distribution (OOD) dataset of hyper-local news synthesized via open-source LLMs (Llama 3, Mistral, Gemma). To analyze model decision boundaries, we apply a tripartite Explainable AI (xAI) framework using SHAP. Our analysis reveals three key phenomena: First, detectors often rely on a "Call-to-Action" syndrome inherent to instruction-tuned LLMs rather than generic machine text. Second, we observe a "Specificity Bias," where LLMs successfully bypass detection by hallucinating specific named entities and mimicking conversational imperfection. Finally, we illustrate the "Wikipedia Paradox," wherein formal, objective human writing is frequently misclassified as AI due to its low statistical burstiness. These findings demonstrate that purely statistical detection paradigms face systematic limitations in highly structured human domains, emphasizing the critical need for semantically-aware, interpretable detection architectures.
\end{abstract}


\section{Introduction}

\IEEEPARstart{T}{he} rapid proliferation of Large Language Models (LLMs), such as Llama 3, Mistral, and GPT-4, has democratized the generation of coherent, contextually nuanced text. As these models become increasingly integrated into daily applications—ranging from automated journalism to social media content creation—the ability to distinguish between human-authored and AI-generated text has emerged as a critical challenge for maintaining digital trust and academic integrity. 

As discussed by Bender et al. \cite{bender2021stochastic}, the risks of large-scale synthetic text generation include the displacement of authentic discourse. This has led to an arms race between generative capabilities and detection methods \cite{wolff2023attacking}, rooted in long-standing stylometric research \cite{stamatatos2009survey}.

While current detection systems, predominantly based on fine-tuned Transformer architectures like BERT and RoBERTa, achieve high accuracies on their respective training distributions, they frequently exhibit severe performance degradation when confronted with OOD data. This vulnerability is particularly pronounced in hyper-local and short-form contexts, where the statistical variance of the text is inherently constrained. Furthermore, these deep learning models operate as opaque "black boxes". Binary Human vs. AI scores provide no insight into the linguistic features or generative artifacts driving predictions, obscuring whether models detect genuine synthetic patterns or merely overfit superficial domain idiosyncrasies.

To address this critical gap, this paper shifts the focus from purely macroscopic performance metrics toward true interpretability using xAI. We systematically evaluate the robustness of fine-tuned Transformer models against domain shifts and utilize SHapley Additive exPlanations (SHAP) to illuminate their decision boundaries. This study is guided by two research questions:
\begin{itemize}
    \item \textbf{RQ1:} How robust are fine-tuned Transformer detectors against extreme out-of-distribution domain shifts, particularly in hy
per-local, short-form text environments?
    \item \textbf{RQ2:} What specific lexical, structural, and semantic patterns drive model misclassifications in both false positive (human flagged as AI) and false negative (AI flagged as human) scenarios?
\end{itemize}

While recent studies have begun to explore SHAP for the explainability of AI text detectors \cite{mitrovic2023chatgpt}, this study uniquely extends this approach by combining global SHAP signatures with systematically selected false negative and false positive case studies in a hyper-local OOD setting. The main contributions of this paper are threefold:
\begin{itemize}
    \item We curate a hyper-local OOD dataset comprising synthetic news articles generated by Llama 3, Mistral, and Gemma alongside authentic human journalism, incorporating randomized temperature parameters to simulate generative variance.
    \item We evaluate the performance drop of baseline models and highlight the utility of diverse training corpora (artem9k) in maintaining low False Positive Rates (FPRs).
    \item We deploy a tripartite xAI framework to analyze the models' vulnerabilities, identifying the "Call-to-Action" syndrome of instruction-tuned LLMs, the "Specificity Bias" that enables AI to bypass detection, and the "Wikipedia Paradox", illustrating how formal human writing triggers AI detection mechanisms.
\end{itemize}
\enlargethispage{\baselineskip}
To support reproducibility and facilitate future research, the complete source code, the custom hyper-local OOD dataset, and the high-resolution SHAP visualizations are publicly available on GitHub at: \mbox{\url{https://github.com/morloher779/xai-analysis}}.

\enlargethispage{\baselineskip}
\section{Related Work}

\subsection{AI Text Detection and Domain Shifts}
The detection of machine-generated text has evolved from traditional statistical approaches to advanced neural methodologies. Baseline models heavily rely on fine-tuning pre-trained Transformers \cite{devlin2018bert}, leveraging the contextual embeddings of models like Bidirectional Encoder Representations from Transformers (BERT) or its robustly optimized variant, RoBERTa \cite{liu2019roberta}, as well as its distilled version, DistilBERT \cite{sanh2019distilbert}, to perform binary sequence classification. However, recent studies highlight the generalization challenges of these systems \cite{mitchell2023detectgpt}. Detectors trained on specific domains, such as academic essays or structured question-answering datasets like HC3 PLUS \cite{su2023hc3plus}, often experience performance degradation when exposed to OOD data, showing variations in recall or increased False Positive Rates.

Recent parallel work by Mady et al. \cite{mady2026feature} has explored feature-augmented Transformers to enhance robustness across diverse domains and generators, utilizing handcrafted linguistic signals to stabilize performance under shift. Our work complements this quantitative mitigation by providing a qualitative xAI analysis, exposing exactly \textit{why} standard architectures fail before such augmentations are applied.
\enlargethispage{\baselineskip}
\subsection{Generative Variance and Dataset Curation}
To counter the issue of domain overfitting, the curation of highly diverse training corpora has become paramount. Datasets like artem9k \cite{artem9k2024aitext} attempt to bridge this gap by aggregating human and AI texts across various platforms, including Reddit and web text. Additionally, the detection landscape is challenged by the generative variance introduced by different sampling parameters. Variations in the temperature setting of LLMs directly manipulate the text's perplexity and burstiness. Consequently, robust evaluation methodologies must simulate this variance to accurately reflect real-world adversarial settings.
\enlargethispage{\baselineskip}
\subsection{Explainable AI (xAI) in Natural Language Processing (NLP)}
Despite advancements in detection accuracy, the interpretability of Transformer-based classifiers remains a challenge. Explainable AI frameworks, particularly SHAP \cite{lundberg2017shap}, have been adapted for NLP to compute token-level attributions, highlighting which words drive a specific classification. While previous research utilizes SHAP primarily for local keyword verification, our work extends this paradigm. Instead of isolated local explanations, we utilize aggregated global SHAP signatures and targeted illustrative case studies (False Negatives and False Positives) to systematically analyze the semantic artifacts and structural biases.
\enlargethispage{\baselineskip}

\section{Methodology}
In this section, we detail the end-to-end pipeline of our research: from the curation of raw datasets and the synthetic generation of out-of-distribution texts, over the architectural fine-tuning, to the core of our study—the Explainable AI (xAI) framework utilized to decode the models' decision boundaries.
\enlargethispage{\baselineskip}
\subsection{Dataset Curation and Synthetic Generation}
To establish a baseline for AI text detection, we fine-tuned our pre-trained Transformer models on two distinct datasets (architectural details are given in Section III.C). The first one is the HC3 PLUS dataset from Su et al. \cite{su2023hc3plus}, which extends the original HC3's human–ChatGPT QA pairs (\textit{GPT-3.5-Turbo}) with semantic-invariant transformations such as paraphrasing, summarization, and translation. This extension better reflects post-edited or re-expressed AI text in real-world practice. The second one is the artem9k/ai-text-detection-pile dataset \cite{artem9k2024aitext}, which is more varied and consists of human and AI-generated text from GPT-2, GPT-3, ChatGPT, and GPT-J. The human text includes a variety of Reddit, web text, and human responses from HC3. This makes the dataset significantly broader and versatile compared to HC3 PLUS.

To evaluate models' generalization capabilities and their vulnerability to domain shifts, we created a custom OOD dataset. We utilized the News\_Category\_Dataset\_v3.json from Kaggle \cite{misra2022news}, which contains around 210,000 headlines from HuffPost between the years 2012 and 2022 spanning various categories, from politics and comedy to home \& living. Using these headlines, we synthesized corresponding articles. To achieve this, we deployed the Ollama framework \cite{ollama_2024} on our local hardware infrastructure and instructed three different state-of-the-art open-source models—Llama 3 \cite{llama3_2024}, Mistral \cite{jiang2023mistral}, and Gemma \cite{gemma_2024}—to generate articles. The prompt for each generation was as follows: \textit{'You are a local community reporter writing for a hyper-local neighborhood app. Your tone is informative, grounded, and straightforward. Avoid overly flowery language, complex metaphors, or typical AI filler words (like 'moreover', 'delve', 'tapestry').'}.

To prevent the detection models from overfitting to a specific statistical signature, the temperature parameter $T$ during the generation process was uniformly randomized, such that $T \sim \mathcal{U}(0.4, 0.9)$. This specific range was deliberately chosen to capture a broad spectrum of generation behaviors, ranging from highly deterministic and factual text ($T=0.4$) to more creative, variable, and unpredictable output ($T=0.9$). In total, we generated a pool of 7,635 AI texts (2,545 per model) and scraped the corresponding human articles. 

From this overarching pool, we constructed the final hold-out OOD test dataset consisting of 4,796 articles. The discrepancy between the initial generation pool (7,635 AI texts) and the final test set arises from the constraints of acquiring the human ground truth. Since the News Category Dataset \cite{misra2022news} provides external URLs to original articles, several sources were unreachable due to dead links or paywalls at the time of scraping. To maintain the rigorous 1:3 class imbalance (1,199 human texts and 3,597 AI texts), AI-generated articles were only included if their corresponding human counterpart was successfully retrieved. This ensures a direct, paired comparison between authentic human reporting and LLM-synthesized content based on identical headlines, while preventing the models from learning spurious correlations based on differing headline distributions.
\enlargethispage{\baselineskip}
\subsection{Data Preprocessing and Tokenization}
To ensure high quality data and prevent artifacts from influencing the model predictions, we systematically preprocessed all raw text corpora. This pipeline included the removal of residual HTML tags, special characters, and structural noise, followed by deduplication to prevent data leakage between the training and testing sets.

The cleaned text sequences were dynamically padded and truncated to a maximum length of 512 tokens to comply with the  input constraints of the Transformers. To achieve this we processed the cleaned data with the help of the pre-trained tokenizers by Hugging Face. This was done for DistilBERT and RoBERTa, respectively.

The synthesized out-of-distribution dataset and the scraped human articles were systematically truncated to reflect real-world constraints typical of social media feeds or short-form journalism. We applied sentence-aware truncation at the first terminal punctuation mark (e.g., period, exclamation mark, or question mark) after ~150 words to avoid artifacts (e.g., missing final punctuation) that detectors could exploit as trivial cues.

This ensures that all input sequences remain syntactically complete. Preventing abruptly severed sentences is methodologically crucial; otherwise, the Transformer models might exploit these structural artifacts (e.g., missing end punctuation) as a trivial shortcut for classification, rather than learning the underlying semantic and linguistic differences between human and AI text. By maintaining natural sentence boundaries, this approach more accurately captures post-edited or re-expressed AI-generated text as observed in real-world usage. Furthermore, enforcing this strict maximum length makes it harder for the detectors to distinguish between human and AI text based purely on length, inherently restricting the evaluable statistical variance (such as burstiness and perplexity) and serving as a deliberate stress test for the models.
\enlargethispage{\baselineskip}
\subsection{Model Architectures and Fine-Tuning Strategy}
To establish a computationally efficient baseline for sequence classification, DistilBERT was employed. DistilBERT is based on BERT, whereby it uses knowledge distillation to retain 97\,\% of BERT’s language understanding capabilities while reducing the parameter count by 40\,\%. This makes it highly suitable for rapid baseline evaluations on the HC3 dataset. Alongside DistilBERT we implemented an advanced architecture for domain-shift analysis. To investigate whether architectural depth mitigates vulnerability to out-of-distribution texts, RoBERTa (roberta-base) was selected as the secondary model. Unlike DistilBERT, RoBERTa utilizes dynamic masking and was pre-trained on a substantially larger corpus, theoretically equipping it with more robust contextual representations necessary for detecting hyper-local, LLM-generated nuances.

Both models were fine-tuned using the Hugging Face Trainer API for binary sequence classification (Human vs. AI). To adapt the pre-trained language models for the specific downstream task of AI text detection, a sequence classification head was appended to the architecture. Specifically, a linear dropout layer followed by a dense projection layer was placed on top of the final hidden state of the special classification token. Note that the exact token notation reflects the models' respective tokenizer vocabularies: \texttt{[CLS]} for DistilBERT and \texttt{<s>} for RoBERTa. This configuration maps the high-dimensional contextual text embeddings directly into a binary label space (Label 0: Human, Label 1: AI).

The end-to-end fine-tuning process was orchestrated utilizing the Hugging Face Transformers library \cite{wolf2020transformers} and its highly optimized Trainer API, ensuring a standardized, hardware-accelerated, and reproducible training loop. The networks were trained by minimizing the standard Cross-Entropy Loss.

The gradient descent optimization was driven by the AdamW algorithm \cite{loshchilov2019decoupled}. In standard Adam, an L2 penalty added to the loss is divided by the adaptive second-moment term, meaning weights with a large gradient history are regularized less \cite{loshchilov2019decoupled}. AdamW addresses this by applying the decay directly to the weights, decoupled from the adaptive step \cite{loshchilov2019decoupled}. This algorithmic adaptation is the standard choice for fine-tuning Transformers \cite{loshchilov2019decoupled}.

To ensure full reproducibility, we detail the fine-tuning hyperparameters: Both models were trained for 3 epochs using a batch size of 16 to accommodate the maximum sequence length of 512 tokens. We employed an initial learning rate of $2 \times 10^{-5}$ with a linear learning rate scheduler and no warmup steps. Early stopping was configured across all training runs by monitoring the validation loss with a patience of 2 epochs, ensuring that the checkpoint with the best validation loss was automatically restored at the end of the training phase.

The selection of hyperparameters followed established empirical best practices for fine-tuning Transformer architectures on specialized downstream tasks \cite{devlin2018bert}. A batch size of 16 was chosen to maintain a balance between memory efficiency and gradient stability on the RTX Pro 5000 Blackwell GPU. The learning rate of $2 \times 10^{-5}$ was selected as it represents the recommended "sweet spot" for BERT-based models, facilitating convergence while preventing the catastrophic forgetting of pre-trained linguistic weights \cite{devlin2018bert}. Furthermore, training was limited to 3 epochs; preliminary tests indicated that additional iterations primarily led to overfitting on the stylistic nuances of the training corpora (e.g., HC3's specific QA-style) rather than enhancing the general detection of artificial signatures.

During inference, alongside the default classification threshold of 0.5, we report the True Positive Rate (TPR) at fixed low False Positive Rates (1\,\% and 5\,\%), reflecting standard operating points in high-stakes environments. Furthermore, to cleanly separate class inseparability from threshold misalignment in a deployment-realistic protocol, we evaluate thresholds calibrated exclusively on the in-distribution validation split and subsequently frozen, thereby avoiding the use of test labels as an oracle.
\enlargethispage{\baselineskip}

\subsection{Explainable AI (xAI) Framework}
While standard performance metrics such as accuracy, recall, and F1-score provide a macroscopic view of model efficacy, they fail to illuminate the underlying decision-making processes of deep Transformer networks. To address this "black-box" phenomenon and shift the research focus toward true interpretability, we integrated the SHAP \textit{PartitionExplainer}, which is specifically optimized for NLP classification pipelines, into our evaluation framework. Based on cooperative game theory, SHAP computes the marginal contribution of each feature—in our context, individual text tokens or sub-tokens—to the final classification output.

Rather than applying xAI solely for isolated local explanations, our methodology leverages SHAP systematically across three distinct analytical paradigms to dissect the models' behavior during severe domain shifts:
\begin{itemize}
   \item \textbf{Global AI Signature Extraction (True Positives):} To identify the universal linguistic patterns learned by the detectors, we aggregate local SHAP values across a representative sample of 50 correctly classified AI texts. To avoid favoring rare tokens, we aggregate subwords to words and report frequency-weighted values based on the number of texts in which a token appears. Furthermore, leveraging the paired structure of our OOD dataset, we compute a contrastive signature (AI texts minus their paired human texts) to cleanly separate LLM stylistic artifacts from underlying topic words.
    \item \textbf{Vulnerability Analysis (False Negatives):} To understand the limitations of statistical detection models, SHAP is applied to synthesized texts that successfully bypassed the detector (hereafter referred to as "Sneaky AI"). This step isolates specific human-mimicking strategies employed by LLMs, allowing us to investigate phenomena such as the specificity bias introduced by hallucinated named entities and rare lexical choices.
    \item \textbf{The Cross-Domain Paradox (False Positives):} To empirically demonstrate the limits of statistical detection, we extend the SHAP analysis to misclassified human texts from a highly formal and structured domain (Wikipedia). This methodological step is designed to visualize how human writing with naturally low statistical variance (low burstiness) triggers the exact same detection mechanisms as AI-generated text.
\end{itemize}
Through this tripartite xAI framework, the methodology transitions from merely benchmarking detection models to uncovering the fundamental lexical vulnerabilities and biases inherent in current AI text detection paradigms.
\enlargethispage{\baselineskip}

\section{Results}
In this section, we present the quantitative performance metrics of the fine-tuned Transformer models. The evaluation focuses on the models' robustness against domain shifts, measuring Accuracy, Macro F1-score, Recall for the AI class (True Positive Rate), and the False Positive Rate (FPR)—the latter being a critical metric for real-world applications where falsely accusing human authors is highly penalized.

\begin{table*}[t]
\centering
\caption{Performance Evaluation of Fine-Tuned Transformer Models Across Different Domains}
\label{tab:model_evaluation}
\begin{tabular}{@{}lllrcccc@{}}
\toprule
\textbf{Model Architecture} & \textbf{Training Data} & \textbf{Test Domain} & \textbf{Accuracy} & \textbf{Precision (AI)} & \textbf{Recall (AI)} & \textbf{F1 (Macro)} & \textbf{FPR (False Positives)} \\ \midrule
DistilBERT    & HC3 PLUS               & OOD (Local News)     & 59.0\%            & \textbf{100.0\%}        & 46.0\%               & 59.0\%              & \textbf{0.08\%} (1/1199)       \\
RoBERTa       & HC3 PLUS               & OOD (Local News)     & \textbf{89.0\%}   & 87.0\%                  & \textbf{100.0\%}     & \textbf{83.0\%}     & 43.37\% (520/1199)             \\
RoBERTa       & artem9k                & OOD (Local News)     & 61.0\%            & \textbf{100.0\%}        & 48.0\%               & 60.0\%              & \textbf{0.08\%} (1/1199)       \\ \midrule
RoBERTa       & artem9k                & Cross-Domain (Wiki)  & 85.0\%            & 84.0\%                  & 87.0\%               & 85.0\%              & 16.95\% (339/2000)             \\ \bottomrule
\end{tabular}

\vspace{1ex}
\begin{minipage}{\textwidth}
\raggedright \footnotesize \textit{Note:} FPR denotes the False Positive Rate. OOD refers to the hyper-local out-of-distribution dataset generated via Llama 3, Mistral, and Gemma.
\end{minipage}
\end{table*}
\enlargethispage{\baselineskip}
\subsection{Performance on Out-of-Distribution Local News}
The initial evaluation was conducted on the custom hyper-local news dataset, representing an extreme OOD scenario characterized by short text lengths (approx. 150 words) and high generative variance. As detailed in Table \ref{tab:model_evaluation}, the computationally lightweight DistilBERT baseline, fine-tuned on the QA-heavy HC3 PLUS dataset, exhibited a highly conservative prediction threshold. It achieved a near-perfect FPR of 0.08\,\% (misclassifying only 1 out of 1,199 human texts) but struggled to detect the LLM-generated texts, resulting in a recall of only 46.0\,\% at the default threshold, despite near-perfect precision. This highlights a severe conservative classification bias rather than statistical inferiority. This gap is primarily attributed to the "QA-bias" inherent in the HC3 PLUS training data. Having been fine-tuned on structured question-answering pairs, the model effectively learned to identify the linguistic markers of a helpful AI assistant (e.g., introductory politeness, structured lists) rather than universal synthetic patterns. When confronted with the objective, reporting-oriented tone of hyper-local news, the model fails to recognize the generative artifacts, as they lack the instruction-following indicators present in its training distribution.

Upgrading the architectural depth to RoBERTa while maintaining the HC3 PLUS training data resulted in overfitting to the target domain. Although the model correctly identified nearly 100\,\% of the AI-generated texts, the FPR surged to 43.37\,\%. This indicates a severe lack of generalization, rendering the model practically unusable in environments where human text is prevalent.

However, training the RoBERTa architecture on the highly diverse artem9k dataset successfully mitigated this domain-shift vulnerability. The model restored the optimal FPR of 0.08\,\% while slightly improving the baseline's AI recall to 48.0\,\% and maintaining an overall accuracy of 61.0\,\%.
\enlargethispage{\baselineskip}
\subsection{Cross-Domain Performance on Formal Text}
To further assess the generalization capabilities and the impact of text length and formal structure on the detection mechanisms, the best-performing model (RoBERTa trained on artem9k) was subjected to a cross-domain evaluation using the GPT-wiki-intro dataset \cite{bhat2024gptwiki}. It is important to acknowledge a methodological asymmetry in this setup: while the primary OOD testing set consists of texts generated by open-source LLMs (Llama 3, Mistral, Gemma), the synthetic portions of this Wikipedia dataset were generated by OpenAI's GPT-3 Curie (\textit{text-curie-001}, temperature 0.7). These texts were synthesized using a 200-word prompt combined with a shared seven-word prefix from the authentic human introductions. 

By simultaneously shifting two variables—the topical domain and the generative architecture—it becomes difficult to isolate whether subsequent performance deviations are driven by the structural formality of Wikipedia or the specific artifacts of GPT-3 Curie. While we explicitly address this confounding factor as a limitation in Section VI, this setup serves as a broad, worst-case evaluation of the detector's overarching robustness in the wild.

The results demonstrate a significant leap in overarching performance metrics compared to the short local news snippets. The model achieved an accuracy of 85.0\,\% and an AI recall of 87.0\,\%, successfully identifying the vast majority of ChatGPT-generated encyclopedia entries. Concurrently, however, the FPR increased significantly to 16.95\,\%, with 339 out of 2,000 genuine human Wikipedia authors being falsely flagged as AI. This stark contrast in FPR between the hyper-local news domain and the academic Wikipedia domain highlights a critical behavioral shift in the detector when confronted with highly structured, formal human writing.
\enlargethispage{\baselineskip}
\subsection{Confusion Matrices and Threshold Dynamics}
While overarching metrics such as Macro F1-score and Accuracy provide a general performance overview, the real-world applicability of AI text detectors is best evaluated through the exact distribution of classification errors. Table \ref{tab:confusion_matrices} provides the detailed Confusion Matrices for all tested configurations, illuminating the specific trade-offs between false accusations (False Positives) and missed synthetic texts (False Negatives).

\begin{table}[htbp]
\centering
\caption{Confusion Matrices for Evaluated Model Configurations}
\label{tab:confusion_matrices}
\resizebox{\columnwidth}{!}{%
\begin{tabular}{@{}llcc@{}}
\toprule
\textbf{Model (Training)} & \textbf{Prediction} & \textbf{True Human} & \textbf{True AI} \\ \midrule
\multirow{2}{*}{\textbf{DistilBERT (HC3)}} & Predicted Human & \textbf{1198} (TN) & 1950 (FN) \\
                                           & Predicted AI    & 1 (FP)             & \textbf{1647} (TP) \\ \midrule
\multirow{2}{*}{\textbf{RoBERTa (HC3)}}    & Predicted Human & \textbf{679} (TN)  & 10 (FN)   \\
                                           & Predicted AI    & 520 (FP)           & \textbf{3587} (TP) \\ \midrule
\multirow{2}{*}{\textbf{RoBERTa (artem9k)}}& Predicted Human & \textbf{1198} (TN) & 1877 (FN) \\
                                           & Predicted AI    & 1 (FP)             & \textbf{1720} (TP) \\ \midrule
\multirow{2}{*}{\begin{tabular}[c]{@{}l@{}}\textbf{RoBERTa (artem9k)}\\ \textbf{Cross-Domain (Wiki)}\end{tabular}} 
                                           & Predicted Human & \textbf{1661} (TN) & 256 (FN)  \\
                                           & Predicted AI    & 339 (FP)           & \textbf{1744} (TP) \\ \bottomrule
\end{tabular}%
}

\vspace{0.3cm}
{\raggedright \footnotesize \textit{Note:} TN = True Negative, FN = False Negative, FP = False Positive, TP = True Positive. Evaluated at a fixed default threshold of 0.5. \par}
\end{table}
\enlargethispage{\baselineskip}
The matrices clearly visualize the stark behavioral differences driven by the training corpora. The DistilBERT baseline and RoBERTa trained on artem9k share a nearly identical, highly conservative profile on the hyper-local OOD dataset. Both models prioritize the protection of human authors at all costs, resulting in a single False Positive out of 1,199 human texts. However, this safety comes at the severe expense of detection capability, letting approximately 52\,\% of the AI-generated texts slip through undetected. This reflects a significant conservative classification bias. Both models prioritize high precision (100\,\%) to minimize the risk of false accusations—a critical requirement for maintaining trust in journalistic or academic contexts. Consequently, in the subtle and neutral domain of local news, the detectors default to a "Human" prediction unless overt, stereotypical AI artifacts are identified. This indicates that the default 0.5 decision boundary is improperly calibrated for short-form OOD detection, as the models' internal confidence scores for AI text are systematically lower in this novel domain.

Conversely, the RoBERTa model trained solely on HC3 PLUS flips this behavior entirely. It acts as an aggressive filter, capturing 3,587 out of 3,597 AI texts (missing only 10), but causing 520 false accusations against genuine human writers. This highlights a fundamental thresholding dilemma: at the default 0.5 decision boundary, current models either act as overly permissive filters or overly aggressive accusers, emphasizing the need for application-specific calibration (e.g., tuning the threshold via validation sets or utilizing AUROC to optimize the operating point) rather than relying on static binary outputs.
\enlargethispage{\baselineskip}
\subsection{Threshold-Independent Evaluation: AUROC and AUPRC}
Given the severe class imbalance in the OOD dataset (1:3 ratio of human to AI texts) and the extreme variations observed in the confusion matrices at the default 0.5 classification threshold, a threshold-independent evaluation is required. To assess the models' inherent ability to separate human from synthetic text across all possible decision boundaries, we computed the Area Under the Receiver Operating Characteristic curve (AUROC) and the Area Under the Precision-Recall Curve (AUPRC). AUPRC is particularly highly informative in this context, as it is more sensitive to false positives in imbalanced datasets than AUROC.

\begin{table}[htbp]
\centering
\caption{Threshold-Independent Performance Metrics}
\label{tab:auroc_metrics}
\begin{tabular}{@{}llcc@{}}
\toprule
\textbf{Model} & \textbf{Test Domain} & \textbf{AUROC} & \textbf{AUPRC} \\ \midrule
DistilBERT (HC3)    & OOD (Local News)     & 0.896          & 0.882          \\
RoBERTa (HC3)       & OOD (Local News)     & 0.806          & 0.841          \\
RoBERTa (artem9k)   & OOD (Local News)     & \textbf{0.995} & \textbf{0.965} \\ \midrule
RoBERTa (artem9k)   & Cross-Domain (Wiki)  & 0.908          & 0.927          \\ \bottomrule
\end{tabular}
\vspace{1ex}
\end{table}

\begin{figure}[htbp]
    \centering
    \includegraphics[width=0.48\textwidth]{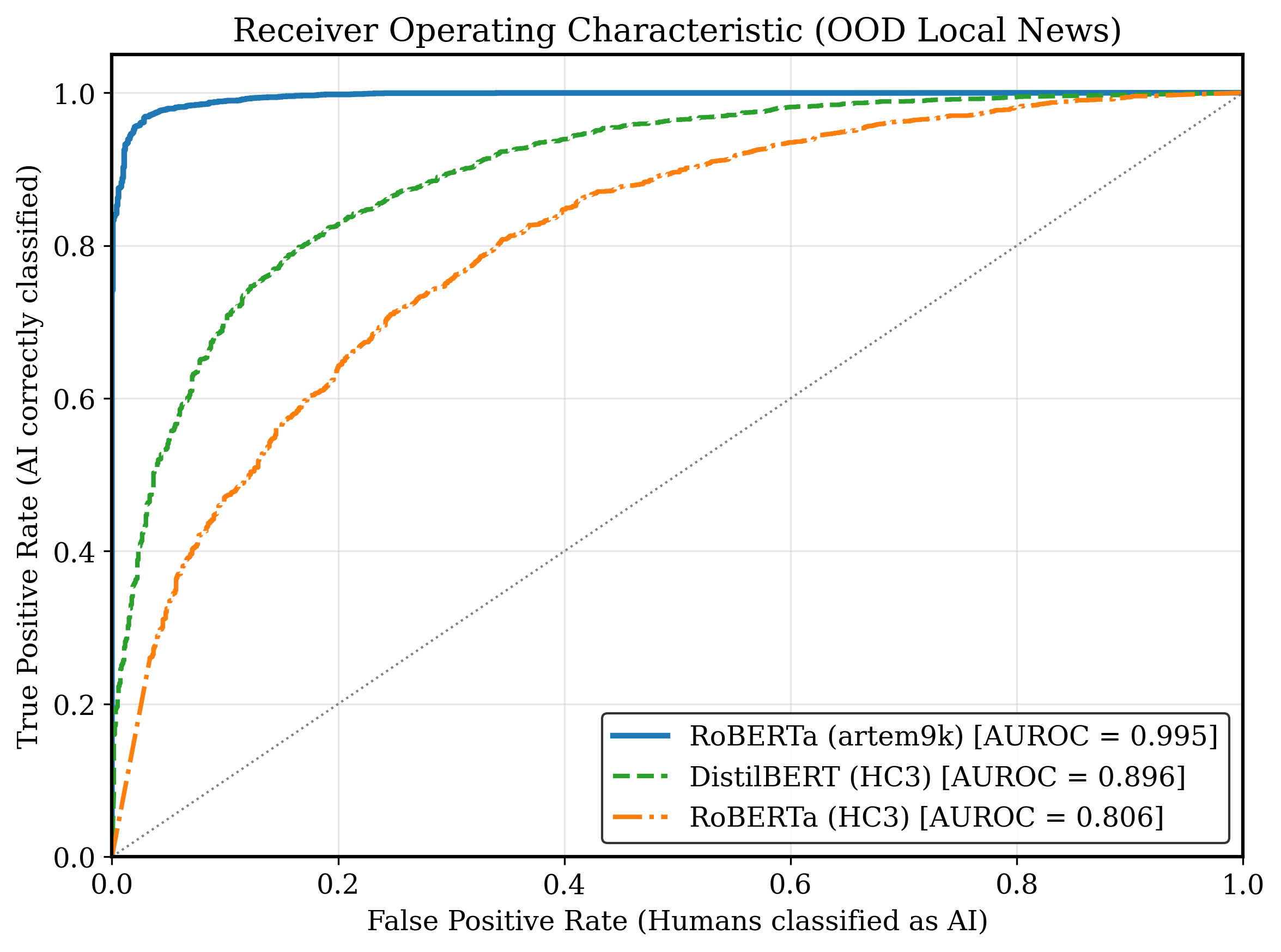}
    \caption{Receiver Operating Characteristic (ROC) curves for the evaluated models on the hyper-local OOD dataset. The curves demonstrate that while RoBERTa (artem9k) achieves excellent class separation, its default 0.5 threshold is severely miscalibrated for this domain.}
    \label{fig:auroc_curves}
\end{figure}

As Table \ref{tab:auroc_metrics} and Figure \ref{fig:auroc_curves} illustrate, the threshold-independent metrics reveal a significantly different performance narrative compared to the static F1-scores previously presented in Table \ref{tab:model_evaluation}. Most notably, RoBERTa trained on artem9k achieves an outstanding AUROC of 0.995 and an AUPRC of 0.965 on the hyper-local OOD dataset. 

This exposes a critical insight regarding model deployment: The underlying contextual embeddings of RoBERTa (artem9k) successfully learn to separate human text from AI text with high confidence. The catastrophic drop in recall (48.0\,\%) observed in Table \ref{tab:model_evaluation} is not a failure of the model's representational power, but rather a severe miscalibration of the default 0.5 decision boundary when transferred to a novel, short-form domain. The model's logits are systematically shifted, requiring a recalibrated threshold to achieve an optimal balance between precision and recall. While tuning the threshold via Youden's J statistic directly on the test domain acts as an oracle ceiling, a threshold fixed strictly on in-distribution validation data offers a deployment-realistic mitigation. Conversely, RoBERTa trained on HC3 PLUS shows a much lower AUROC (0.806), confirming that its high FPR is caused by a fundamental inability to separate the classes in the OOD space, rather than just a thresholding misalignment.
\enlargethispage{\baselineskip}
\section{Discussion}
The quantitative results and confusion matrices highlight a fundamental trade-off in current AI text detection paradigms. From a practical deployment perspective, the configurations exhibit vastly different real-world consequences. While the DistilBERT baseline minimizes false accusations, missing over half of the AI-generated content (46.0\,\% Recall) makes it highly problematic for use cases like automated content moderation, where filtering synthetic spam is the primary goal. Conversely, the 43.37\,\% FPR observed in RoBERTa-HC3 indicates massive overfitting to the target domain. In practical terms, this renders the model entirely unsuitable for high-stakes environments—such as academic grading or journalistic vetting—where falsely accusing a human author of using AI carries severe reputational penalties.

To decode the underlying reasons for these performance variations and to understand how modern LLMs (Llama 3, Mistral, Gemma) mimic or fail to mimic human writing, we analyze the models through our tripartite xAI framework.

In this section we're working with two different colors—red and blue—to visualize the decision making of our chosen transformers. Red symbolizes a push towards AI-creation for any token and blue means the transformer evaluates it as human-made.

\subsection{Vulnerabilities in AI Detection: A SHAP Analysis}

\subsubsection{The Global AI Signature and 'Assistance Behavior'}
To extract the overarching lexical features that the RoBERTa model associates with artificial generation, we computed the global SHAP values across correctly classified AI texts (True Positives). 

\begin{figure}[htbp]
    \centering
    \includegraphics[width=0.48\textwidth]{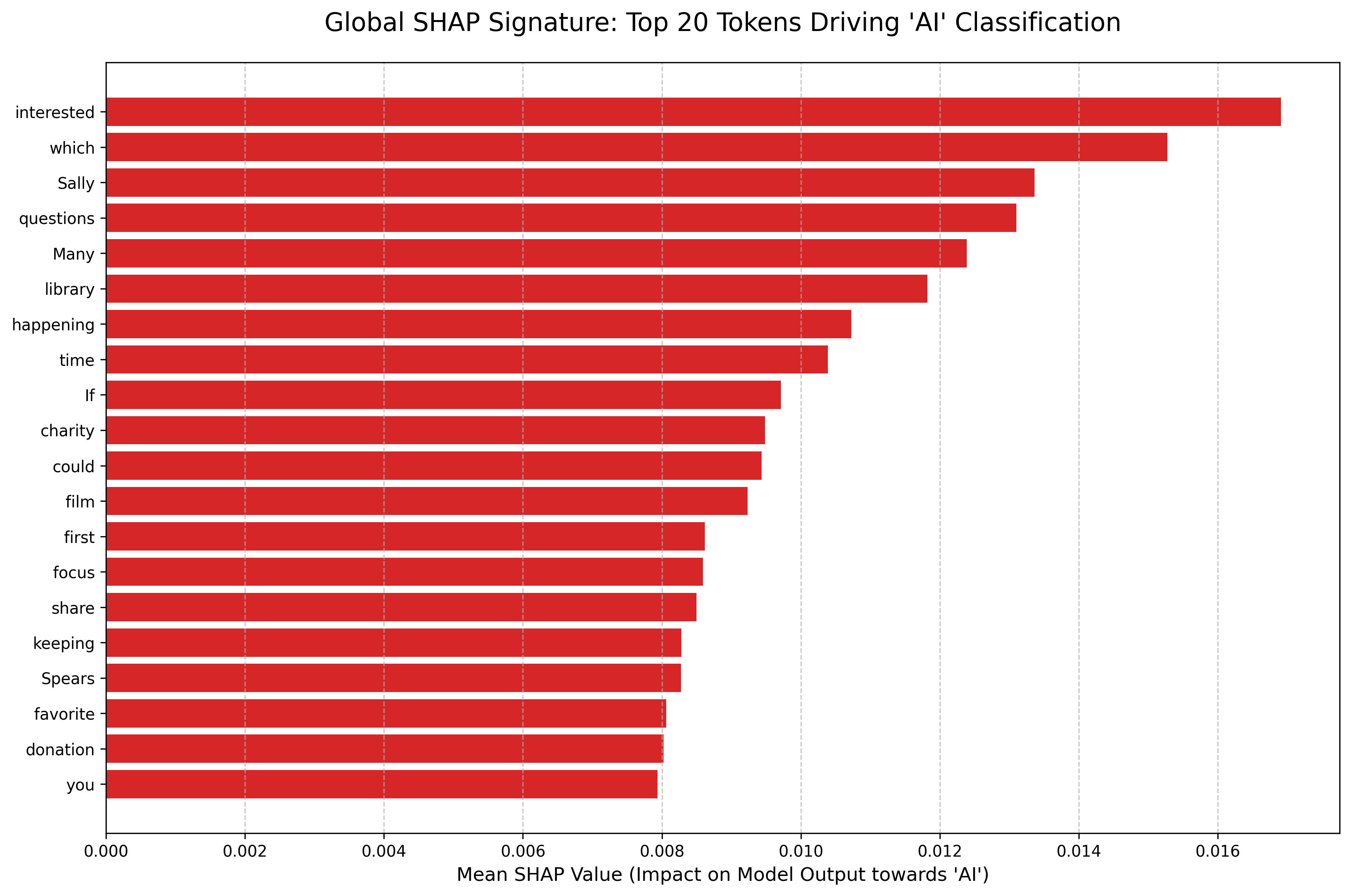}
    \caption{Global SHAP Signature: Top tokens driving the 'AI' classification in hyper-local news.}
    \label{fig:shap_global}
\end{figure}

The global SHAP analysis reveals that the detector did not rely on stereotypical AI filler words (which were explicitly suppressed via the generation prompt). Instead, it identified deeper generative narratives. The highest positive impact values for the AI class were assigned to tokens such as \textit{'If'}, \textit{'interested'}, and \textit{'questions'}. This exposes a profound behavioral artifact of instruction-tuned LLMs: the "Call-to-Action" syndrome. Unlike human journalists who report objectively, models like Llama 3, Mistral or Gemma inherently default to an assistance-oriented tone, frequently concluding articles with service-oriented phrases (e.g., "If you are interested, please reach out"). Furthermore, the prominence of tokens like \textit{'charity'} and \textit{'Sally'} indicates that the models attempt to simulate hyper-local contexts by hallucinating wholesome community stereotypes, a pattern the detector successfully learned to flag.
\enlargethispage{\baselineskip}

\subsubsection{The True Positive: Generic Framing and the "Feel-Good" Bias}
\begin{figure}[htbp]
    \centering
    \includegraphics[width=0.48\textwidth]{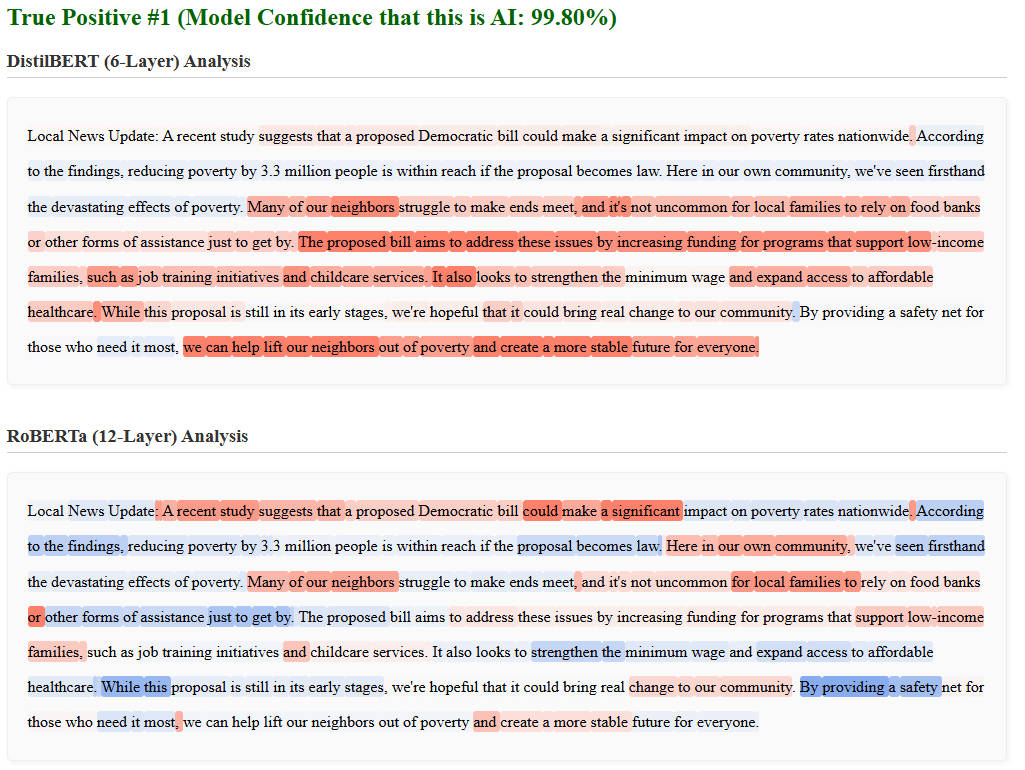}
\caption{Architectural Comparison on a True Positive (RoBERTa (artem9k) Confidence: 99.80\,\%). Both models exhibit $>99\,\%$ certainty of artificial origin. Red tokens push toward AI; blue tokens push toward Human.}
    \label{fig:shap_tp}
\end{figure}

Figure \ref{fig:shap_tp} illustrates an extreme True Positive case—a synthetic article about a poverty bill where the models are over 99.8\,\% confident of its artificial origin. The visualization reveals a striking difference in both architectural depth and training data influence (DistilBERT fine-tuned on HC3 PLUS versus RoBERTa fine-tuned on artem9k). While training DistilBERT on artem9k would complete a 2x2 experimental design to isolate architectural depth perfectly, the current setup confirms that shallow models trained on QA-datasets overfit to the generic, helpful tone typical of instruction-tuned LLMs.

In contrast, RoBERTa exhibits a much more granular understanding. It flags generalized, non-specific journalistic framing in red (e.g., "A recent study suggests", "could make a significant"), but notably highlights the specific statistical claim "reducing poverty by 3.3 million people" in blue. RoBERTa recognizes that concrete numbers are typically human journalistic traits, but the overwhelming presence of synthetic filler words still rightfully triggers the overall AI classification.

\subsubsection{The False Negative: Specificity Bias and Human Imperfection}
\begin{figure}[htbp]
    \centering
    \includegraphics[width=0.48\textwidth]{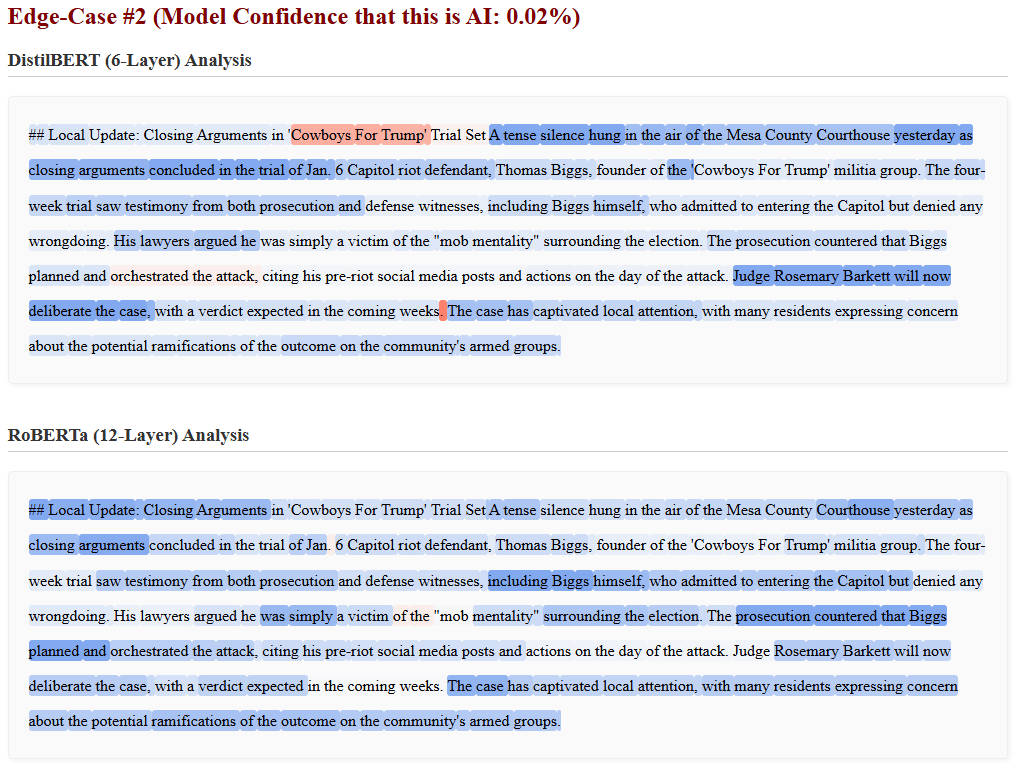}
    \caption{Architectural Comparison on a False Negative ("Sneaky AI", 0.02\,\% AI Confidence, DistilBERT (HC3 PLUS)). The presence of specific Named Entities overwhelmingly pushes the prediction toward Human (blue).}
    \label{fig:shap_fn}
\end{figure}

To understand the critical limitations of these detectors, Figure \ref{fig:shap_fn} shows a "Sneaky AI" False Negative—a synthetic text that tricked both models into near 100\,\% certainty that it was written by a human. The text reports on a fictionalized "Cowboys For Trump" trial.

The SHAP heatmap clearly exposes the \textit{Specificity Bias}. Both DistilBERT and RoBERTa are overwhelmingly pushed toward the 'Human' class (blue) by highly specific, hallucinated Named Entities, such as "Mesa County Courthouse", "Thomas Biggs", and "Judge Rosemary Barkett". Furthermore, the dramatic journalistic opening ("A tense silence hung in the air") is heavily weighted as human. 

\begin{figure}[htbp]
    \centering
    \includegraphics[width=0.48\textwidth]{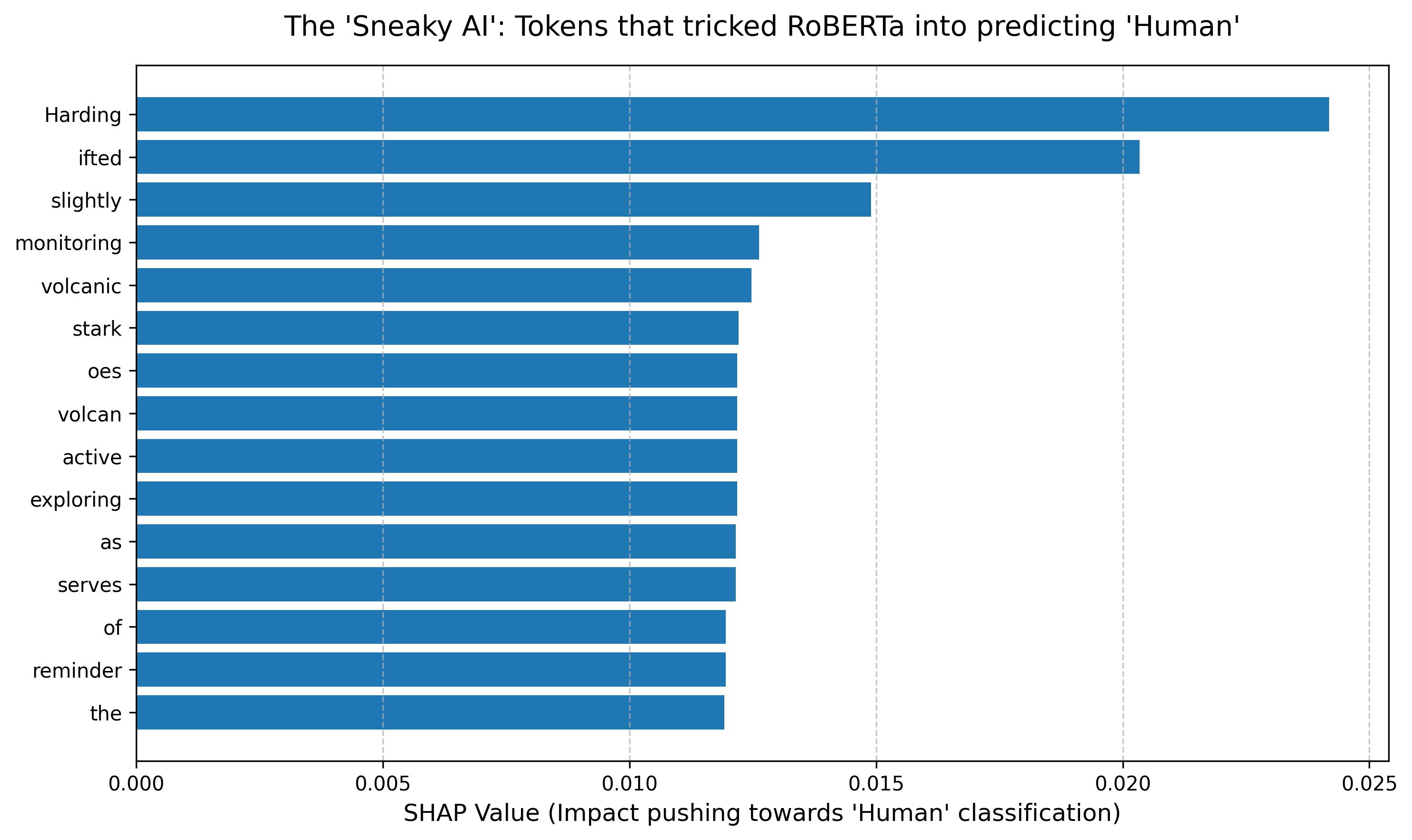}
    \caption{Local SHAP Analysis of a False Negative ('Sneaky AI'): Tokens pushing the prediction toward the 'Human' class.}
    \label{fig:shap_sneaky}
\end{figure}

This observation is further corroborated by the local ranked token attribution for this specific text, shown in Figure \ref{fig:shap_sneaky}. Instead of relying on generic abstractions, the model generated specific Named Entities, such as the fictional tourist "Harding", which had the absolute highest impact score (+0.0242) pushing the prediction toward the human class. In the training corpora, such factual specificity regarding previously unmentioned individuals correlates strongly with human journalism. Furthermore, the high impact of the sub-token "ifted" (derived from "airlifted") shows the LLM utilized rare vocabulary rather than predictable verbs like "transported". The phrase "slightly conscious" introduces a degree of conversational imperfection, mimicking a human eyewitness account.

\enlargethispage{\baselineskip}
This dual visualization illustrates that statistical detectors do not possess a semantic understanding of truth; they merely associate a dense web of specific proper nouns and stylistic narrative flair with human journalism. When an LLM successfully hallucinates these features, it bypasses the detection mechanisms entirely.
\subsubsection{The Wikipedia Paradox: Limits of Statistical Detection}
\begin{figure}[htbp]
    \centering
    \includegraphics[width=0.48\textwidth]{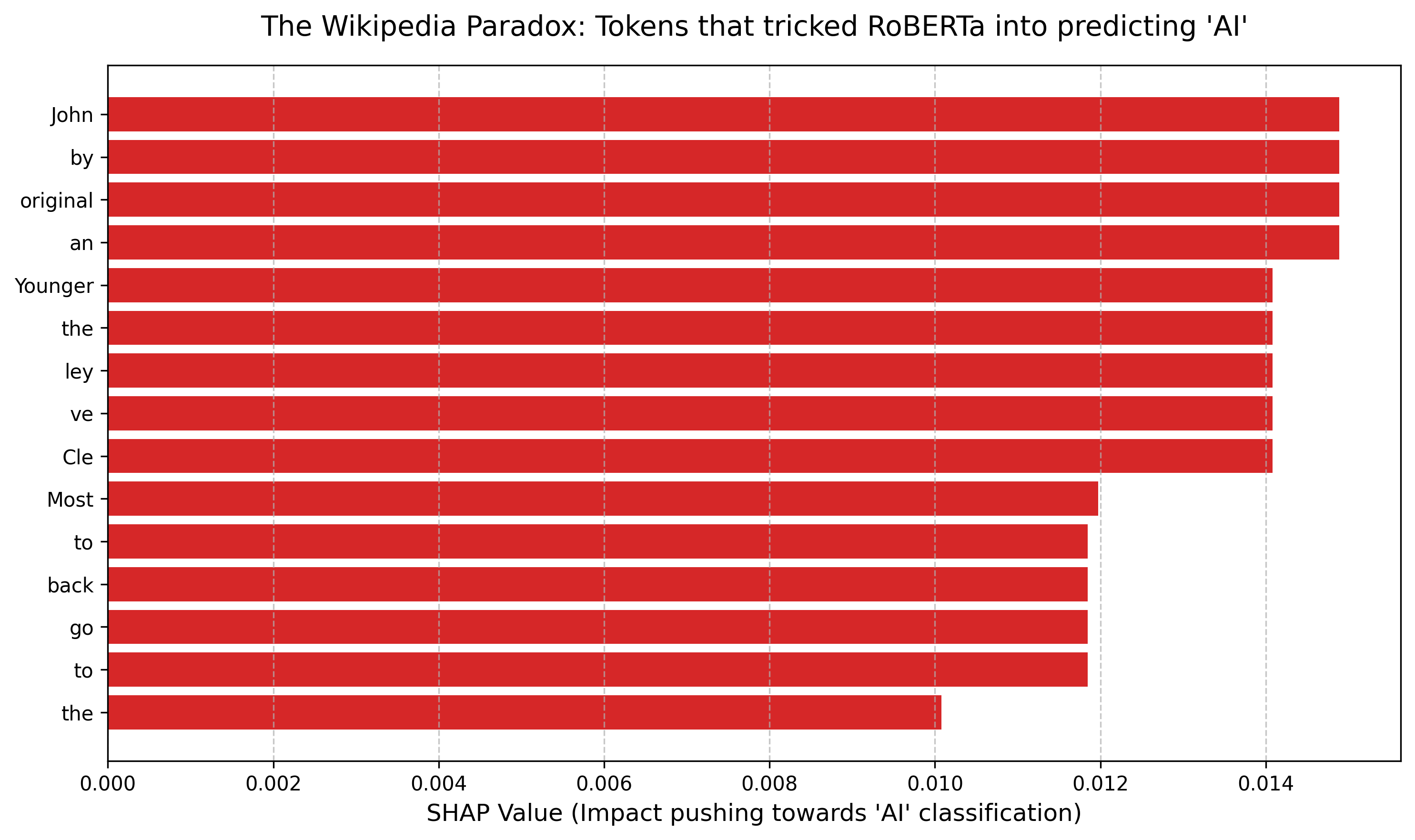}
    \caption{The Wikipedia Paradox: Local SHAP values of a genuine human text misclassified as AI.}
    \label{fig:shap_wiki}
\end{figure}

The cross-domain evaluation on the Wikipedia dataset yielded a significantly higher FPR of 16.95\,\%, exposing a critical limitation of current detection paradigms. To investigate why genuine human authors were frequently flagged as artificial, we extracted the SHAP values of a misclassified, human-written encyclopedia entry (Figure \ref{fig:shap_wiki}).
\enlargethispage{\baselineskip}

The tokens driving the 'AI' prediction include factual entities, structural descriptors, and standard grammatical markers: \textit{'John'} (+0.0149), \textit{'original'} (+0.0149), and \textit{'an'} (+0.0149).This supports the hypothesis that the detector associates structural formality, factual density, and emotional detachment with artificial generation. While burstiness and perplexity were hypothesized but not empirically measured in this study, future work should directly compute these metrics (e.g., via GPT-2) across the 339 misclassified texts versus the correctly classified ones to test the "Wikipedia Paradox" quantitatively. Furthermore, aggregating SHAP attributions over all 339 false positives would solidify the impact of these objective function words beyond isolated case studies.
\enlargethispage{\baselineskip}
\section{Limitations}
While this study provides valuable interpretability insights into the decision-making processes of AI text detectors, several methodological limitations must be acknowledged. First, the out-of-distribution evaluation is constrained to a single hyper-local news domain and Wikipedia introductions, operating exclusively in the English language. 

Second, modern LLMs function as highly adaptive adversaries. Our "Sneaky AI" analysis serves as an initial, illustrative step toward understanding adversarial generation patterns. However, it is not an exhaustive robustness evaluation against deliberate evasion techniques, such as targeted prompt injection, structural obfuscation, or multi-agent rewriting pipelines. 

Third, there is an intentional dataset asymmetry that impacts broad generalization claims: the primary OOD test set focuses exclusively on open-source generation models (Llama 3, Mistral, Gemma), whereas the cross-domain Wikipedia dataset (GPT-wiki-intro) evaluates texts generated by OpenAI's GPT-3 Curie. Additional confounding variables include the reliance on a single training seed, as well as a potential prompt confound between the hyper-local instructions given to the LLMs and the national scope of the scraped human news articles.
\enlargethispage{\baselineskip}
\section{Conclusion}
This study investigated the robustness and interpretability of Transformer-based AI text detectors under domain shifts. By evaluating against a custom OOD dataset generated by open-source LLMs (Llama 3, Mistral, Gemma), we observed that models fine-tuned on diverse corpora (artem9k) generalize better and maintain lower False Positive Rates compared to those trained on narrow datasets (HC3 PLUS). Through the application of a tripartite SHAP framework, we demonstrated that detectors rely on specific generative artifacts, such as the "Call-to-Action" syndrome, rather than generic AI markers. We also illustrated the "Specificity Bias," where LLMs bypass detection via targeted named entity hallucination, and the "Wikipedia Paradox," showing that formal, low-burstiness human writing often triggers false positive classifications. 
\enlargethispage{\baselineskip}
These findings suggest that purely statistical detection paradigms face inherent limitations when applied to formal human domains. Future work should expand this evaluation to multi-lingual contexts to assess cross-linguistic generative patterns. Furthermore, the development of next-generation detectors should explore contrastive training approaches to better separate formal human intellect from synthetic generation, and investigate watermarking-based hybrid detection systems to establish more reliable, semantic-aware authentication mechanisms.

% --- REFERENCES ---
\bibliographystyle{IEEEtran}
\bibliography{references}

\clearpage
\onecolumn 
\appendices

\section{Extended SHAP Analysis: The Wikipedia Paradox}
This section provides additional examples of the ``Wikipedia Paradox'' discussed in Section IV.B. The SHAP values illustrate how formal, structured vocabulary inherently pushes the model toward an artificial classification, resulting in false positives for genuine human encyclopedic texts.

\begin{figure}[H]
    \centering
    \begin{subfigure}{0.42\textwidth}
        \includegraphics[width=\linewidth]{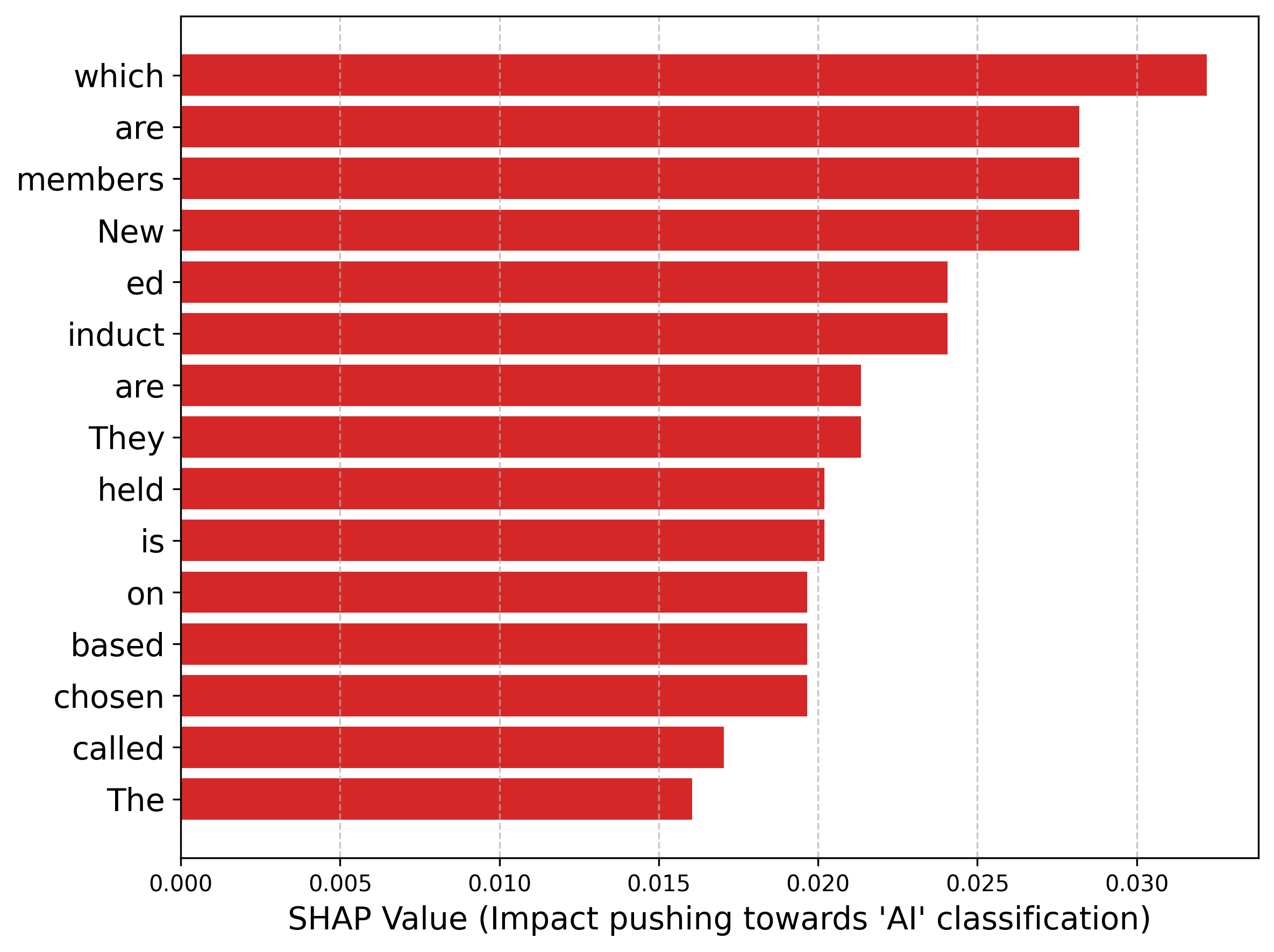}
        \caption{Example 1}
    \end{subfigure}\hfill
    \begin{subfigure}{0.42\textwidth}
        \includegraphics[width=\linewidth]{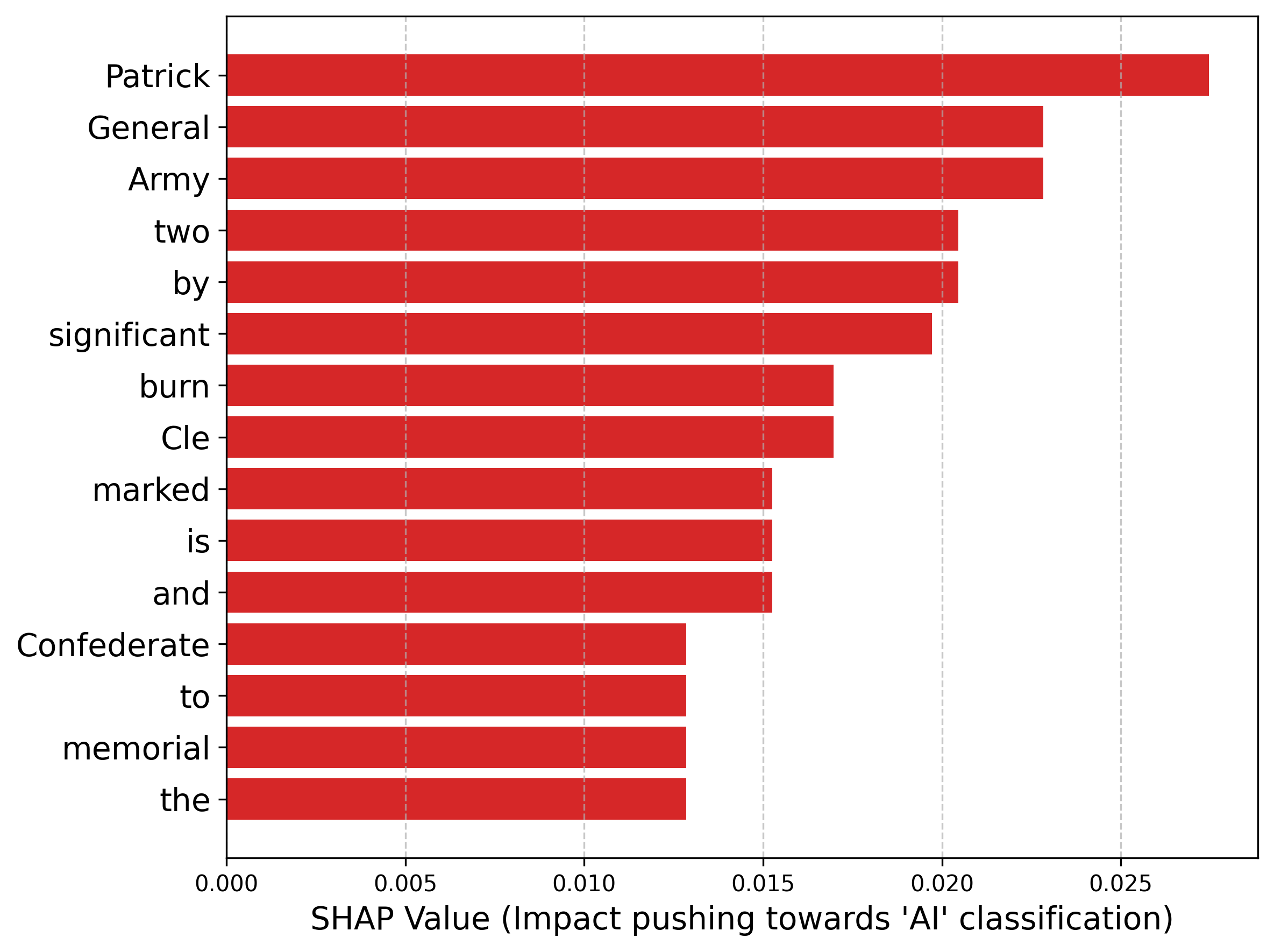}
        \caption{Example 2}
    \end{subfigure}
    
    \vspace{0.1cm}
    
    \begin{subfigure}{0.42\textwidth}
        \includegraphics[width=\linewidth]{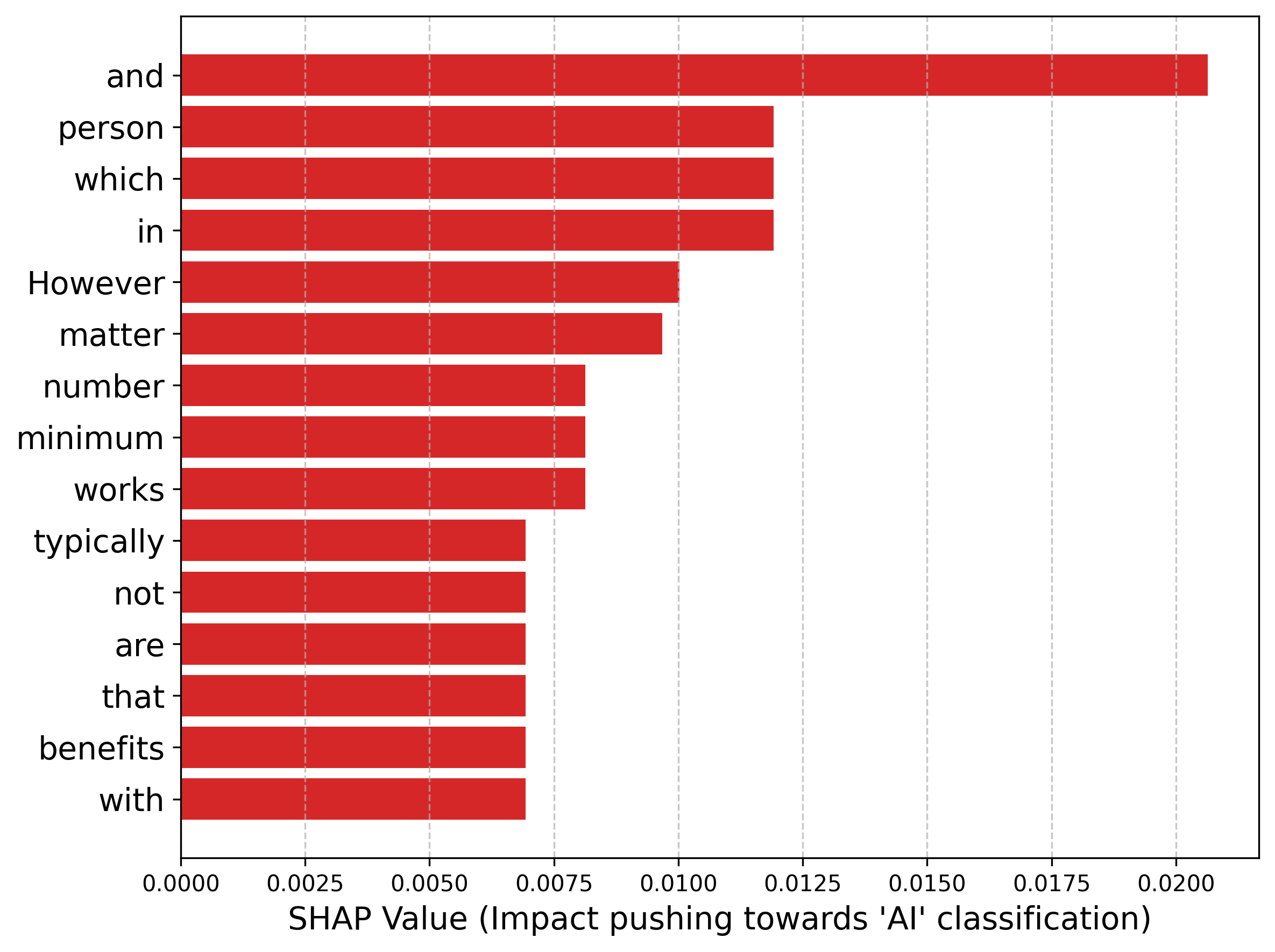}
        \caption{Example 3}
    \end{subfigure}\hfill
    \begin{subfigure}{0.42\textwidth}
        \includegraphics[width=\linewidth]{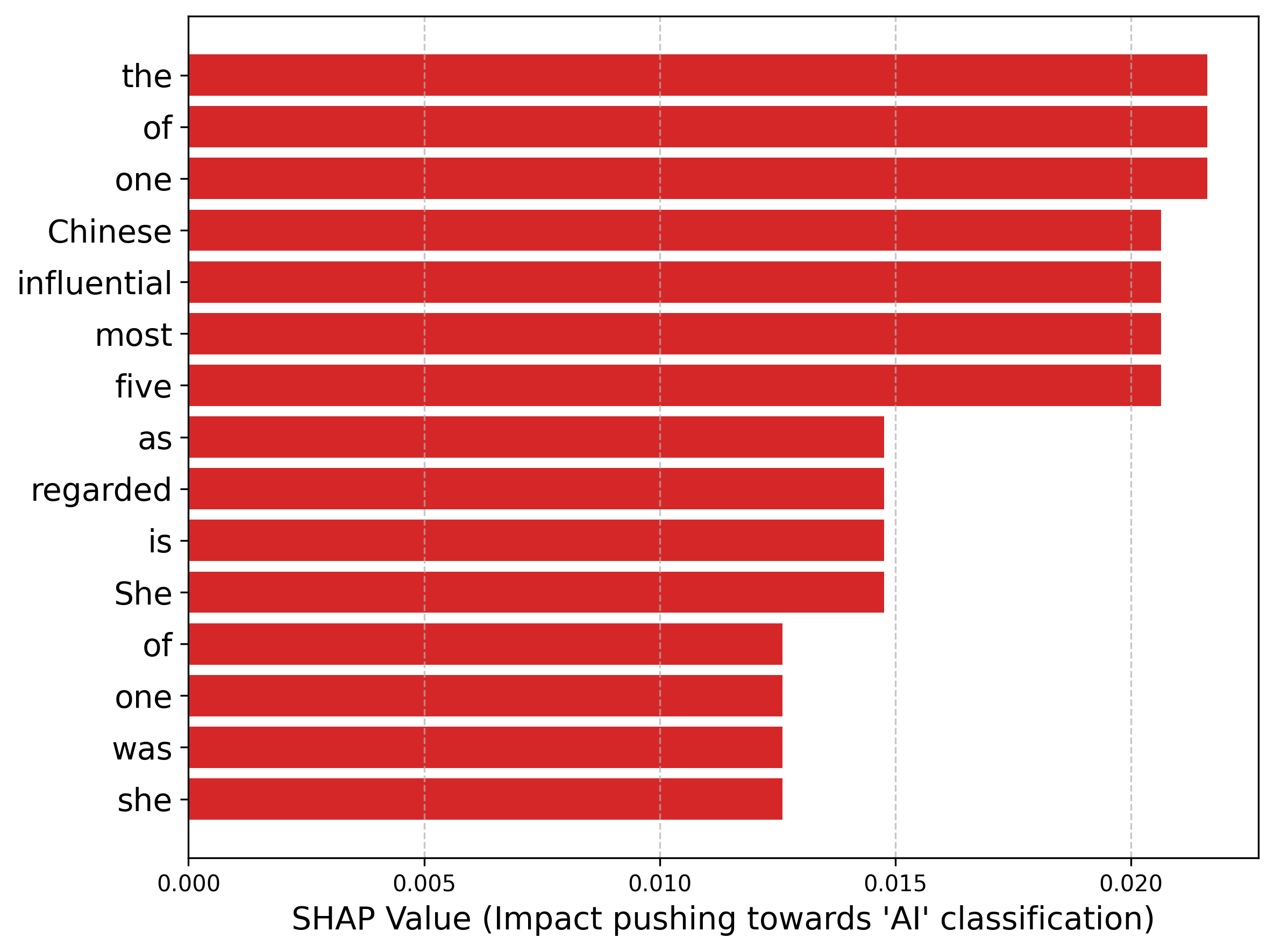}
        \caption{Example 4}
    \end{subfigure}
    
    \vspace{0.1cm}
    
    \begin{subfigure}{0.42\textwidth}
        \includegraphics[width=\linewidth]{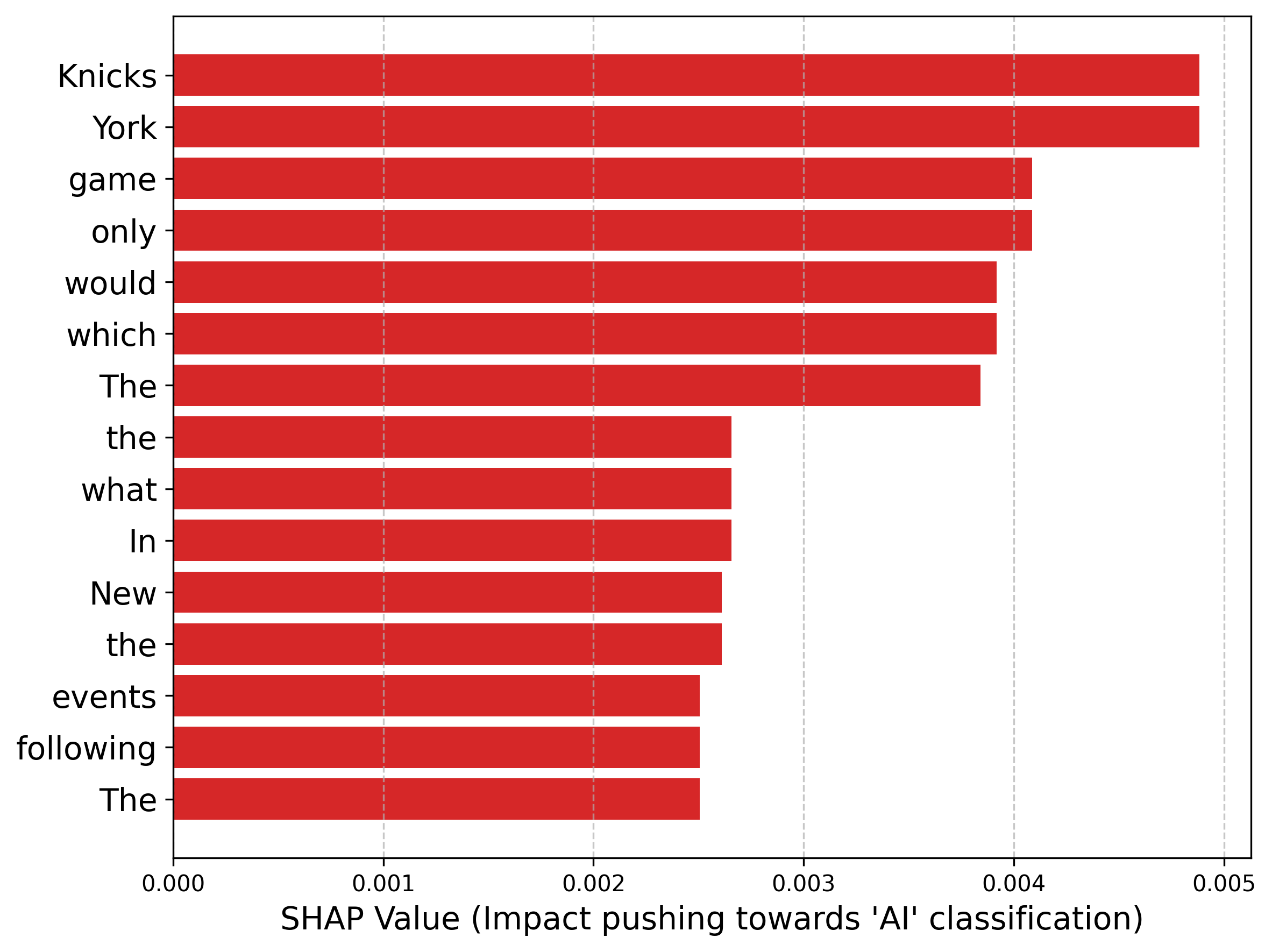}
        \caption{Example 5}
    \end{subfigure}
    \caption{Extended SHAP analysis of the Wikipedia Paradox. Tokens pushing the classification toward 'AI' are highlighted in red.}
    \label{fig:appendix_wiki_grid}
\end{figure}

\clearpage

\section{Extended SHAP Analysis: Cross-Domain Vulnerabilities}
Expanding upon the findings in Section V, this section visualizes additional SHAP attribution profiles for Out-of-Distribution (OOD) local news texts. It highlights both the ``Sneaky AI'' phenomenon (false negatives) and the ``Feel-Good'' bias (true positives).

\begin{figure}[H]
    \centering
    \begin{subfigure}{0.42\textwidth}
        \includegraphics[width=\linewidth]{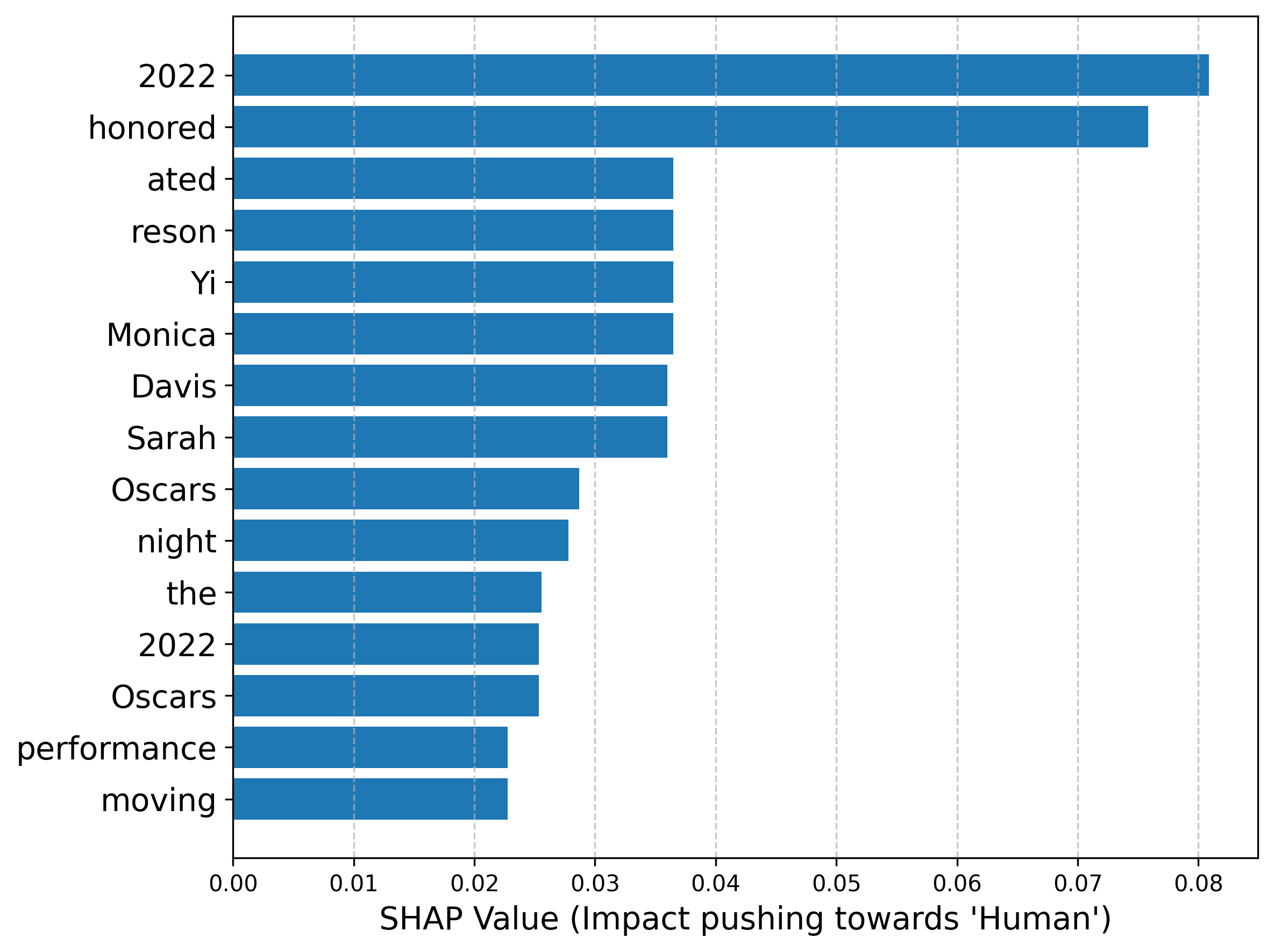}
        \caption{Example 1}
    \end{subfigure}\hfill
    \begin{subfigure}{0.42\textwidth}
        \includegraphics[width=\linewidth]{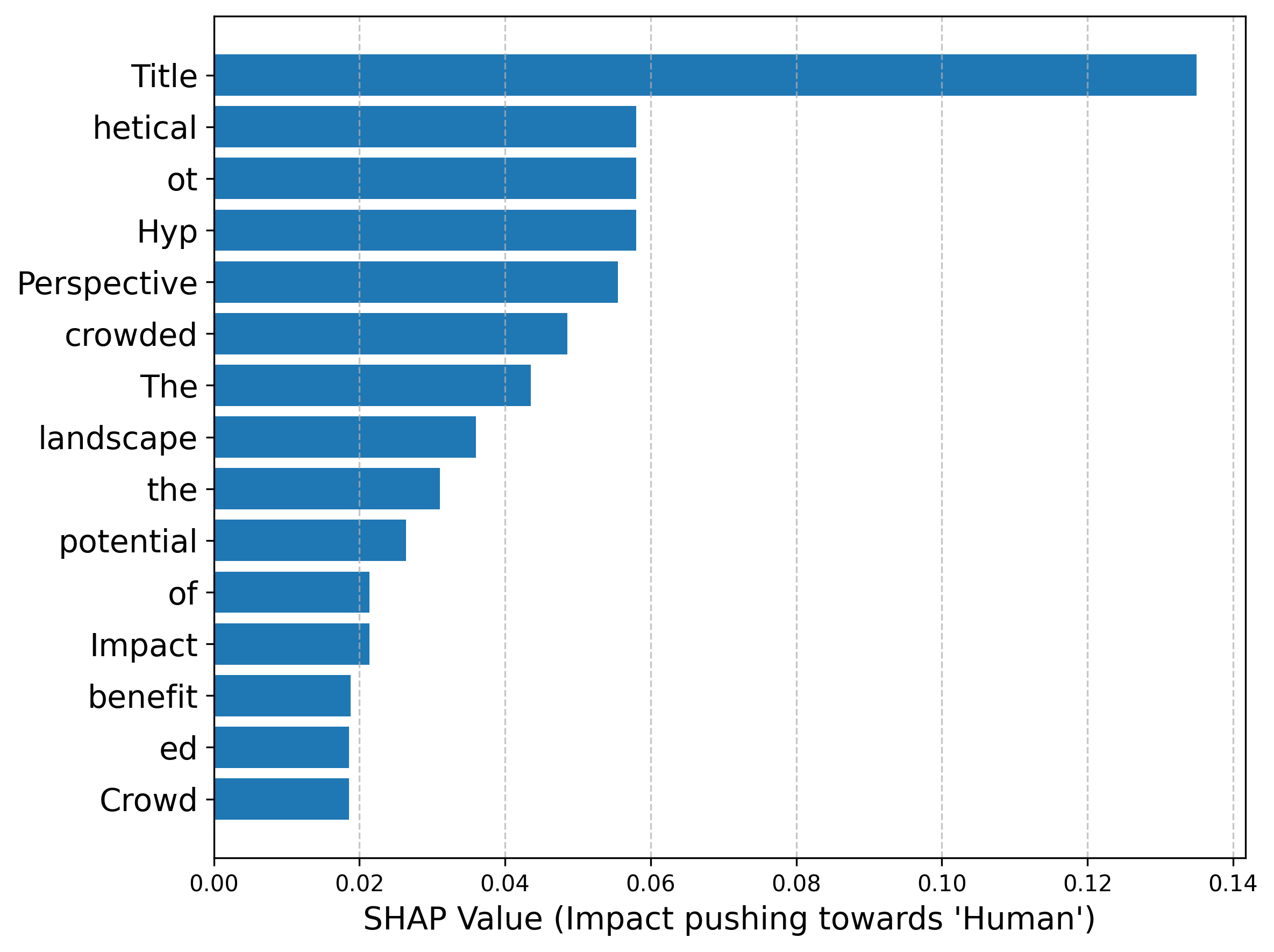}
        \caption{Example 2}
    \end{subfigure}
    
    \vspace{0.1cm}
    
    \begin{subfigure}{0.42\textwidth}
        \includegraphics[width=\linewidth]{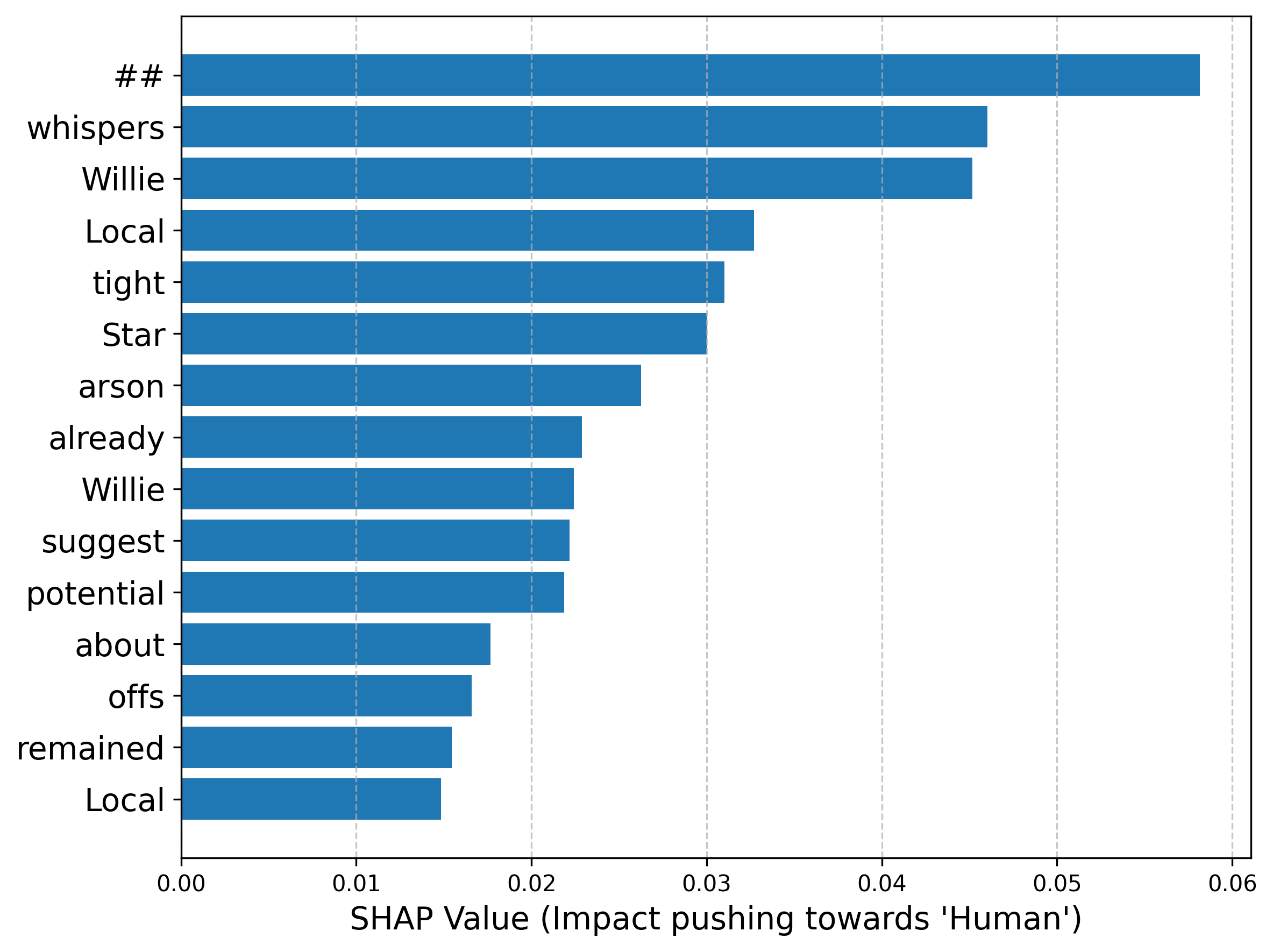}
        \caption{Example 3}
    \end{subfigure}\hfill
    \begin{subfigure}{0.42\textwidth}
        \includegraphics[width=\linewidth]{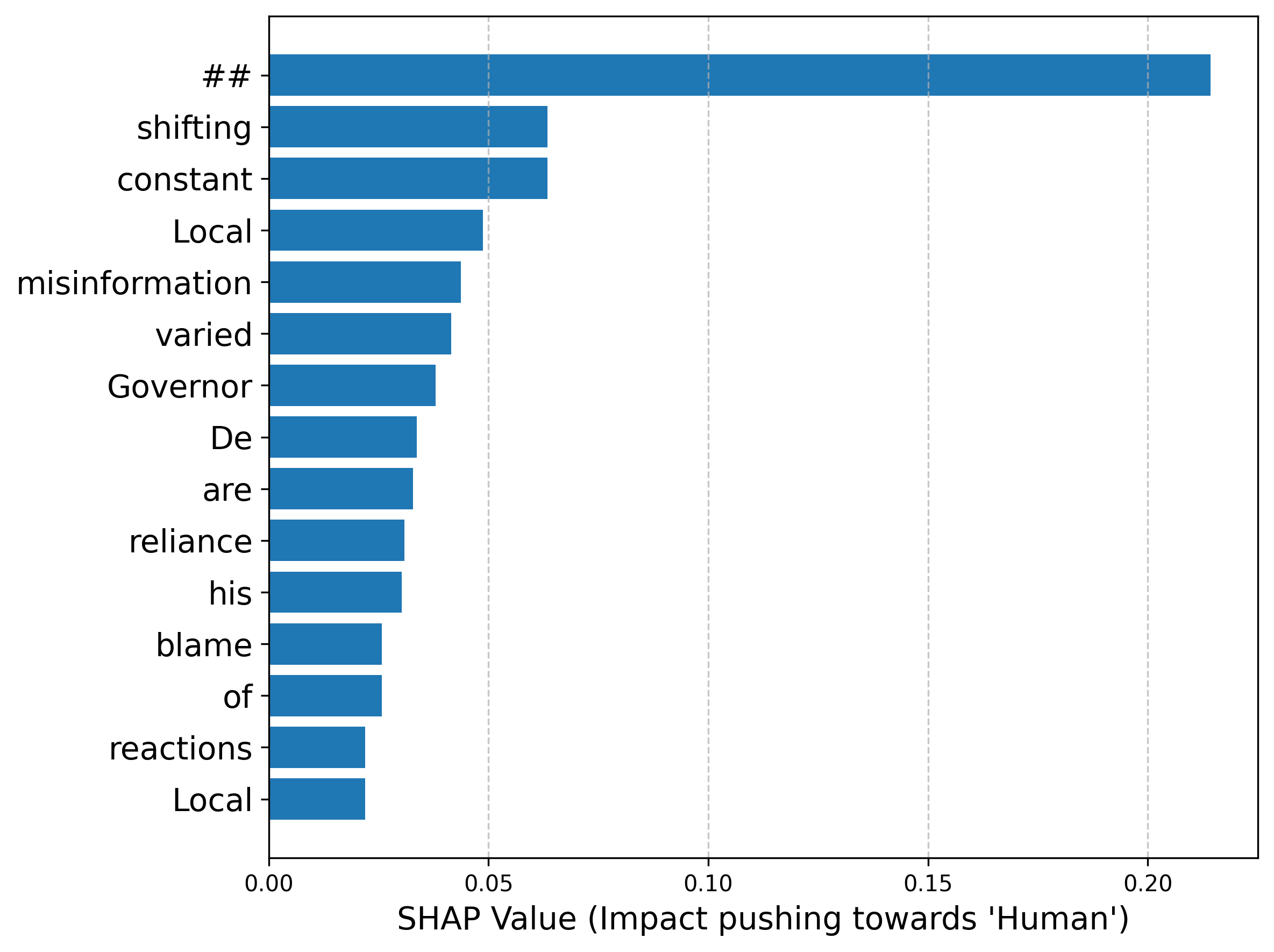}
        \caption{Example 4}
    \end{subfigure}
    
    \vspace{0.1cm}
    
    \begin{subfigure}{0.42\textwidth}
        \includegraphics[width=\linewidth]{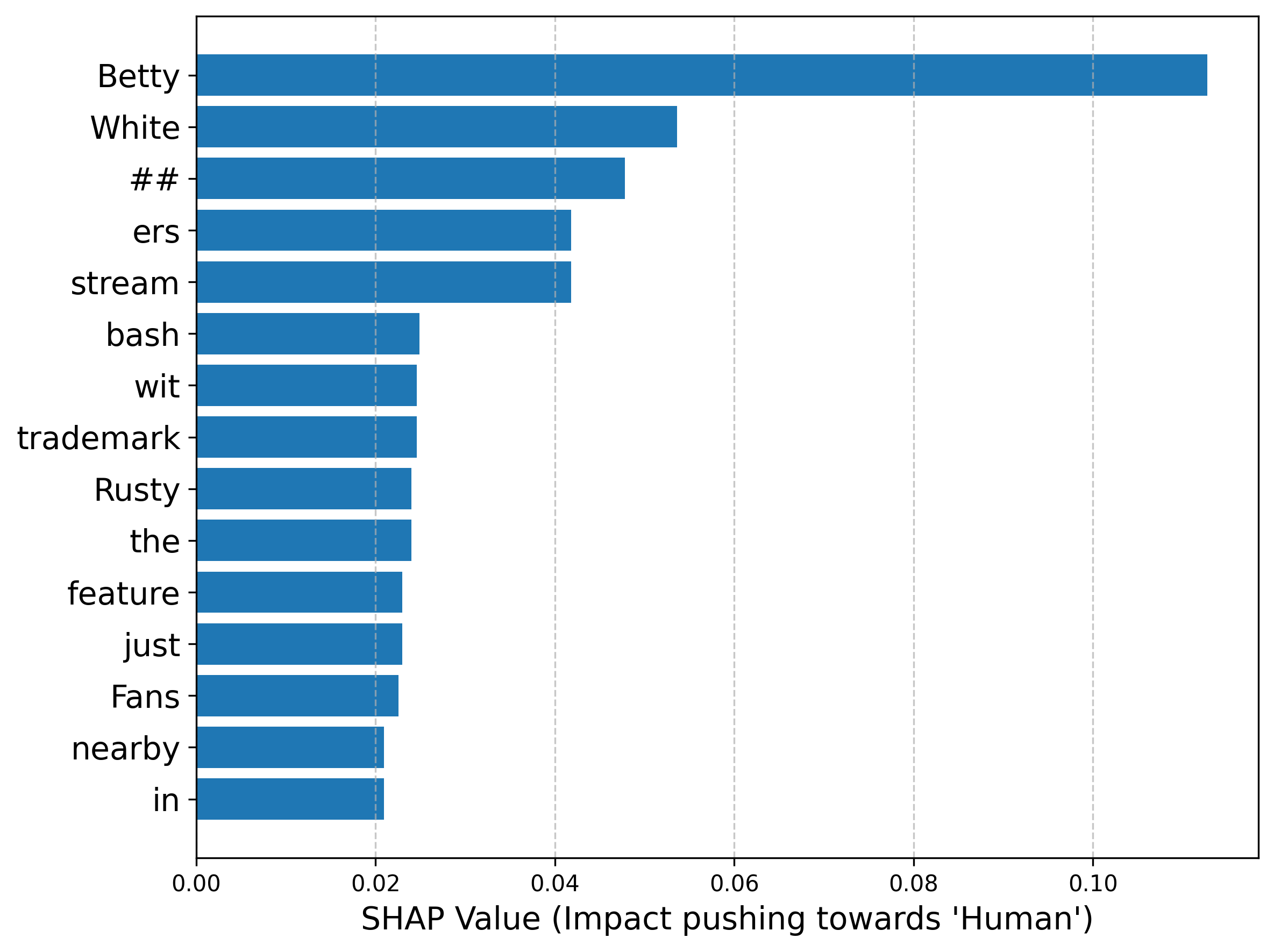}
        \caption{Example 5}
    \end{subfigure}
    \caption{Sneaky AI (False Negatives). Tokens pushing the model toward a 'Human' prediction are highlighted in blue.}
    \label{fig:appendix_sneaky_grid}
\end{figure}

\clearpage

\section{Extended SHAP Analysis: True Positives}
This section illustrates clear instances where the models correctly rely on synthetic text artifacts and generic assistance behavior to classify LLM-generated texts as artificial.

\begin{figure}[H]
    \centering
    \begin{subfigure}{0.42\textwidth}
        \includegraphics[width=\linewidth]{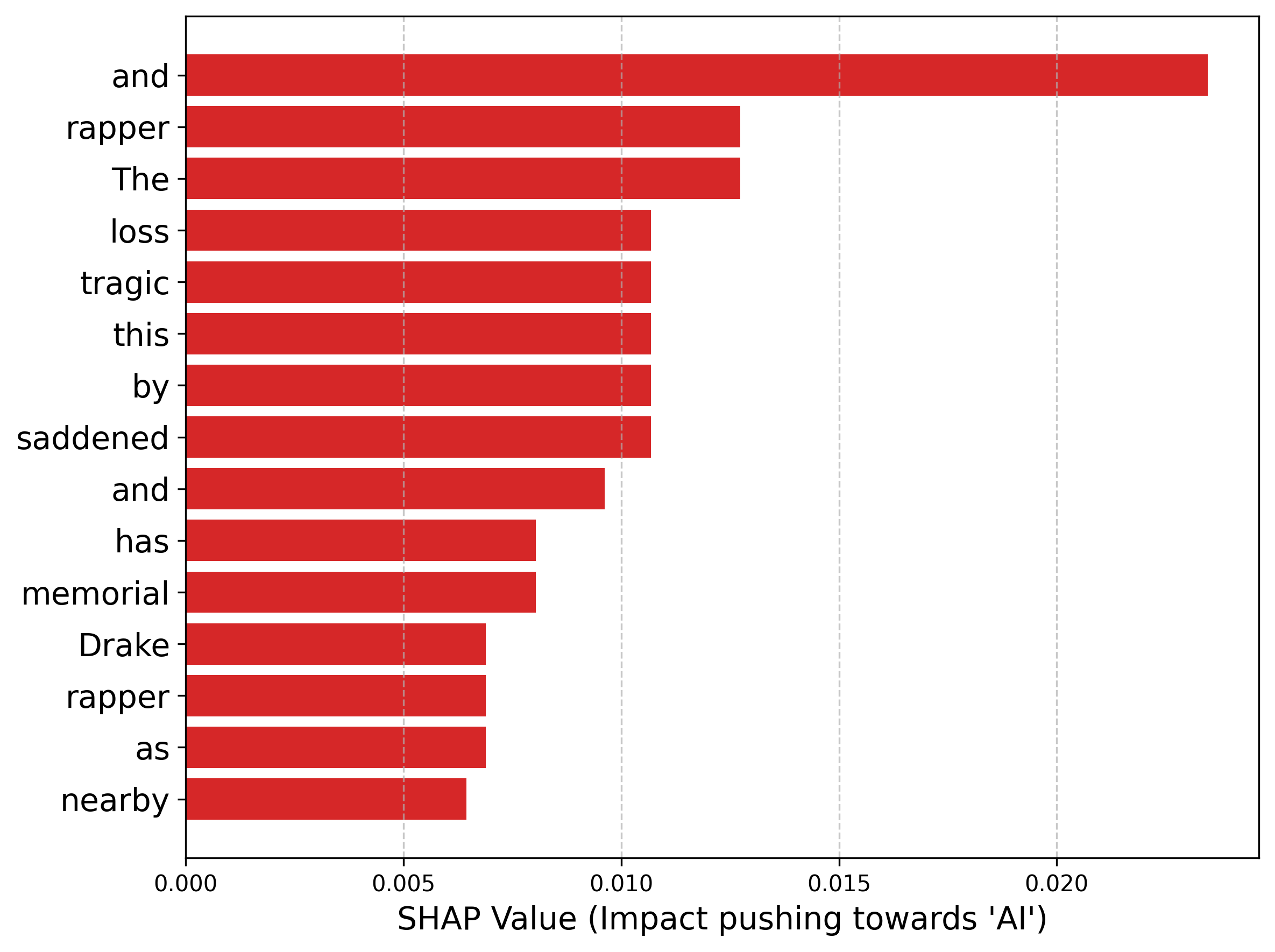}
        \caption{Example 1}
    \end{subfigure}\hfill
    \begin{subfigure}{0.42\textwidth}
        \includegraphics[width=\linewidth]{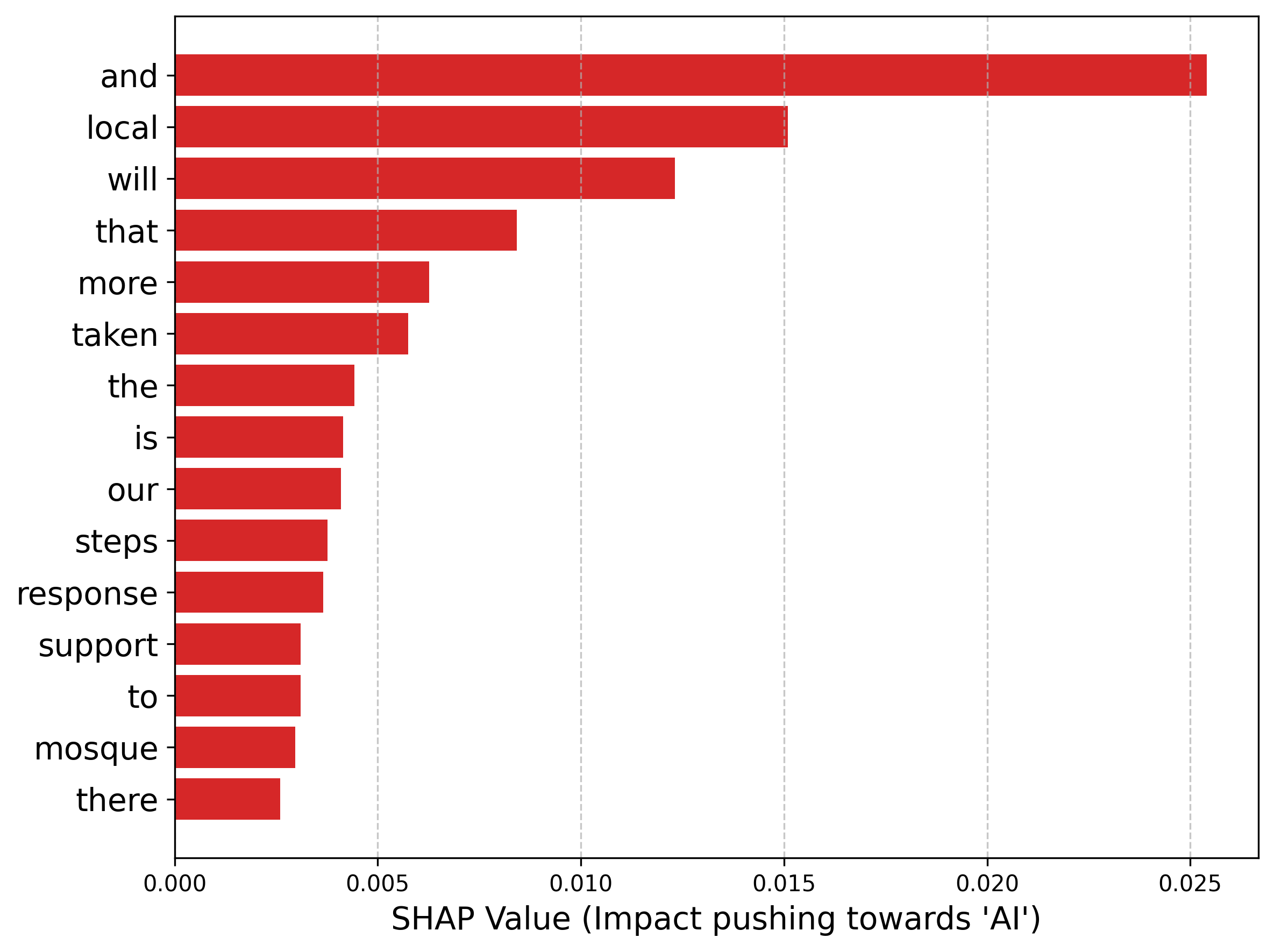}
        \caption{Example 2}
    \end{subfigure}
    
    \vspace{0.1cm}
    
    \begin{subfigure}{0.42\textwidth}
        \includegraphics[width=\linewidth]{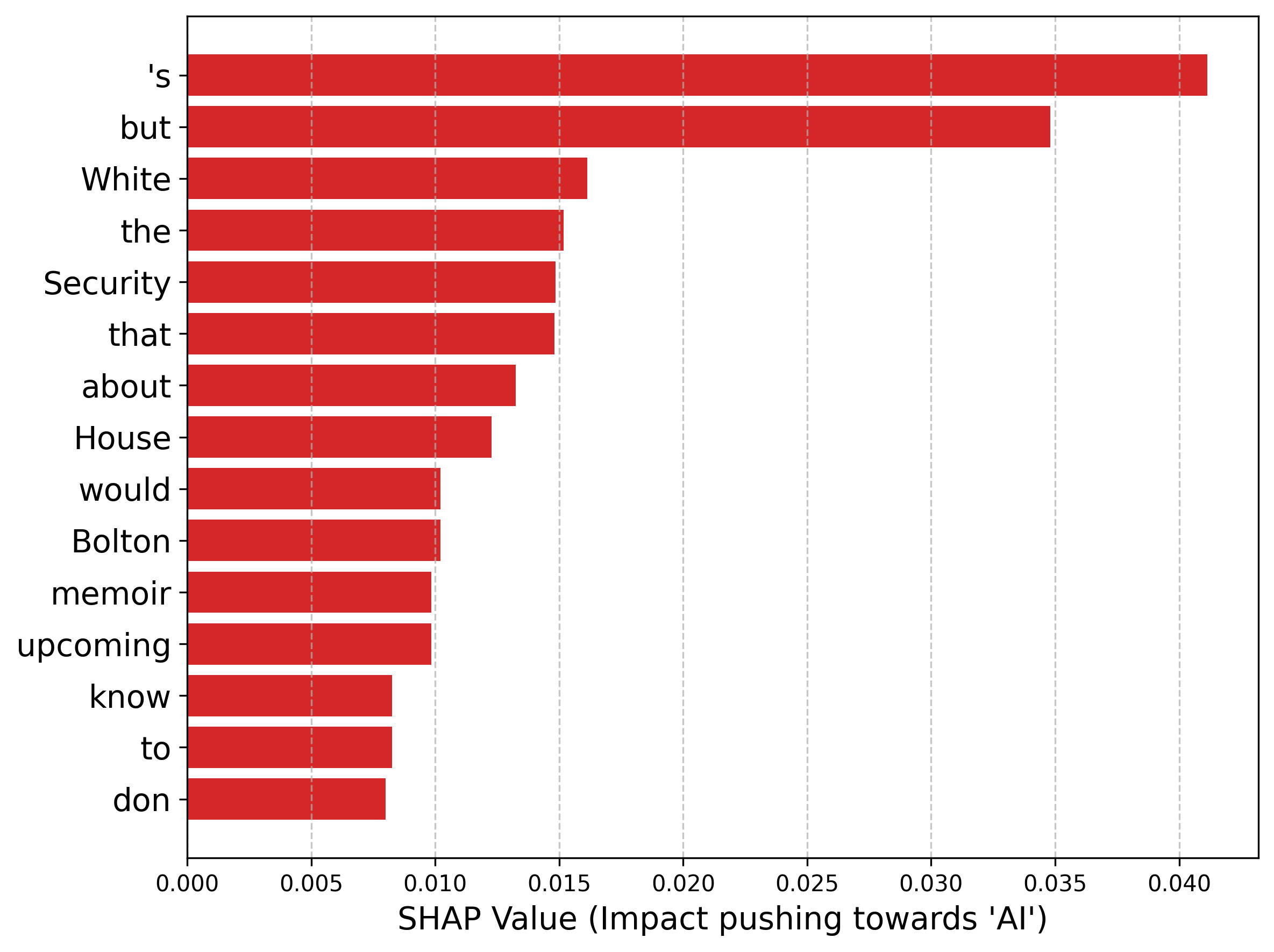}
        \caption{Example 3}
    \end{subfigure}\hfill
    \begin{subfigure}{0.42\textwidth}
        \includegraphics[width=\linewidth]{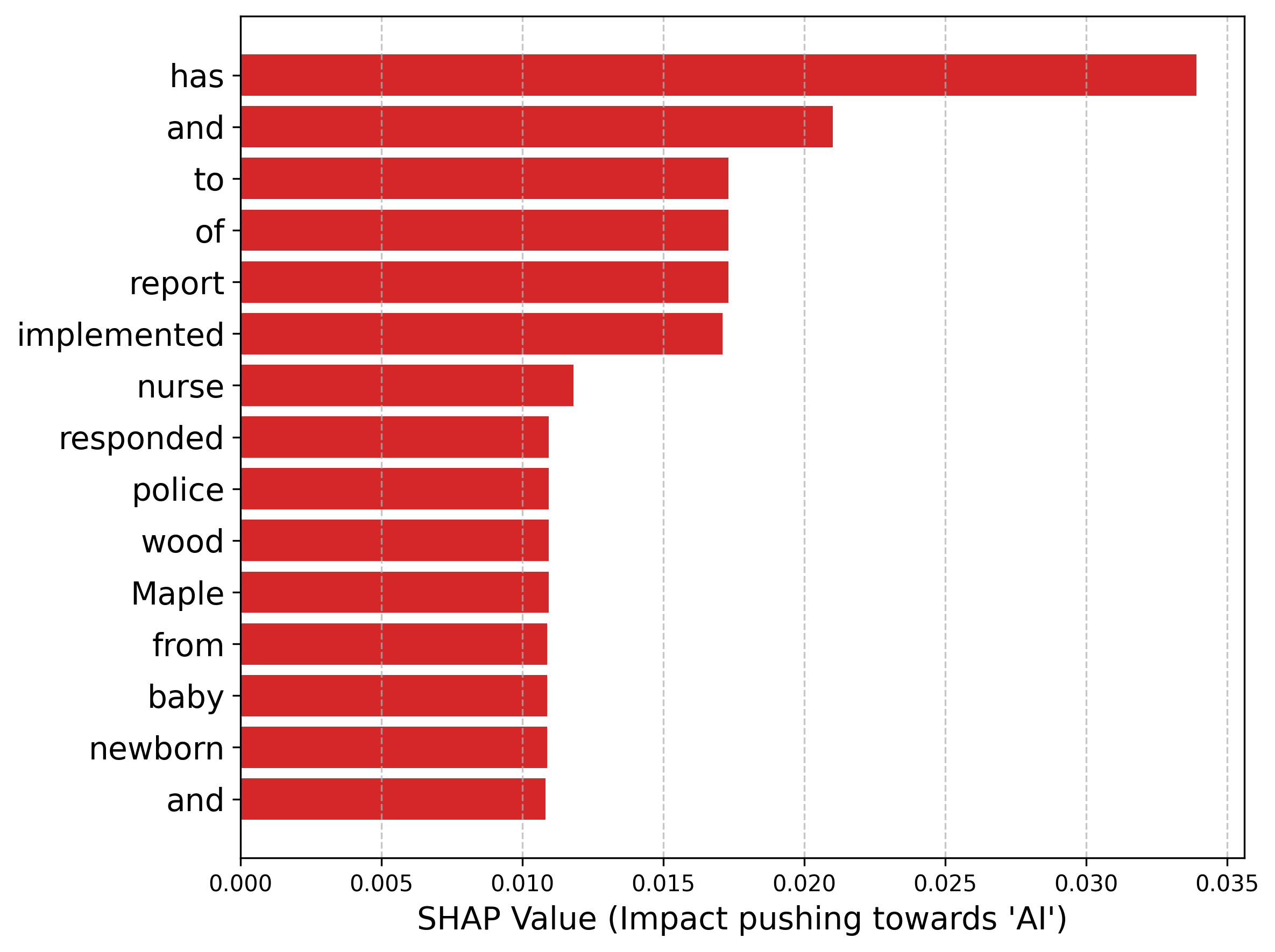}
        \caption{Example 4}
    \end{subfigure}
    
    \vspace{0.1cm}
    
    \begin{subfigure}{0.42\textwidth}
        \includegraphics[width=\linewidth]{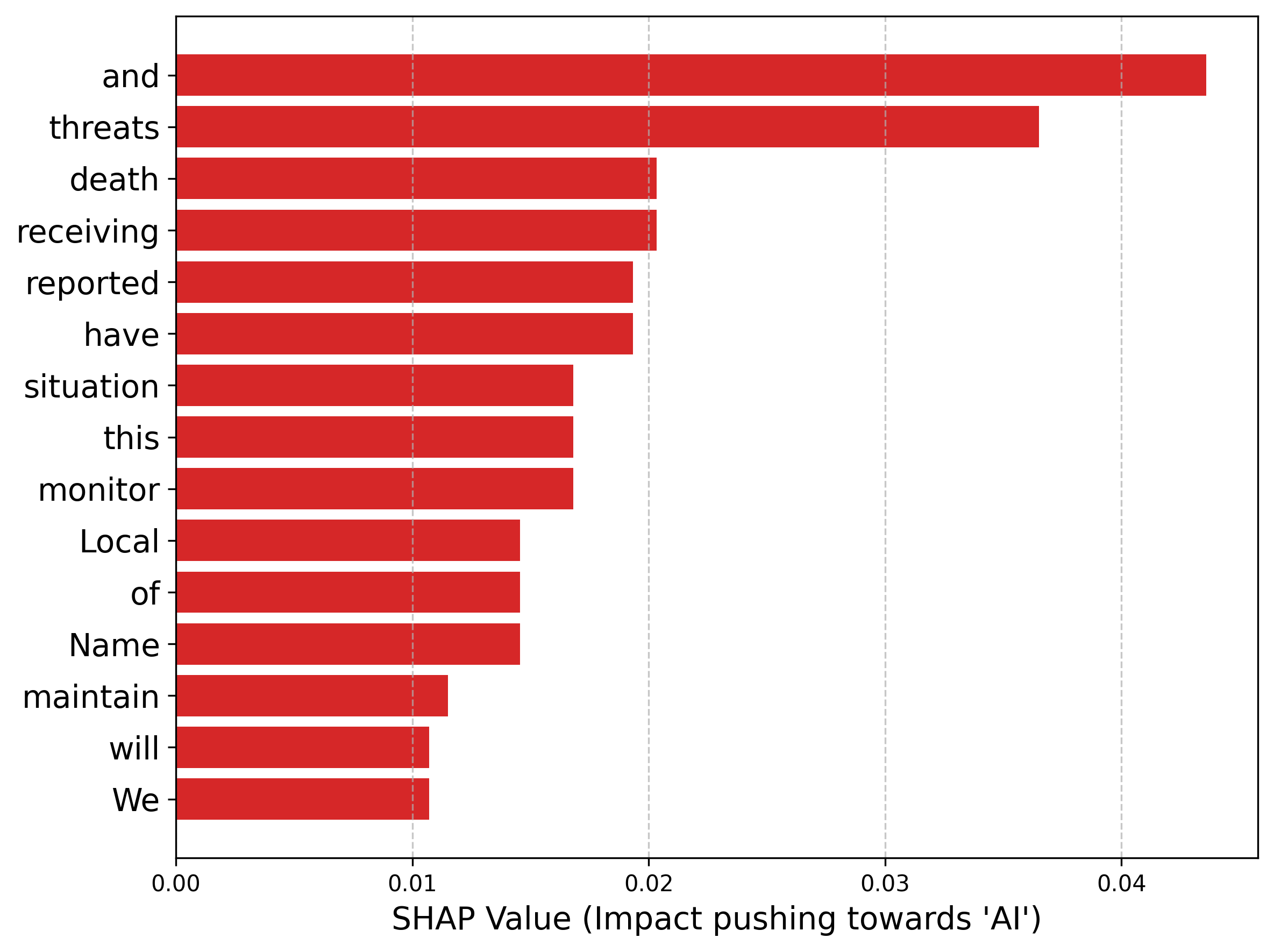}
        \caption{Example 5}
    \end{subfigure}
    \caption{True Positives. Tokens correctly pushing the model toward an 'AI' prediction are highlighted in red.}
    \label{fig:appendix_tp_grid}
\end{figure}

\clearpage
\section{Evaluated Text Transcripts}
This section provides the original 150-word text transcripts corresponding to the SHAP visualizations presented in the previous appendices.

\subsection{The Wikipedia Paradox}
\small % Etwas kleinere Schriftart, um Platz zu sparen
\textbf{Example 1 (corresponding to Figure \ref{fig:appendix_wiki_grid}):} The Raven's Claw is an all-male senior honorary society at Dickinson College. It was founded in 1896, making it the first society unique to Dickinson College and one of the oldest in the country. Membership is limited to seven senior men who are selected by the seven previous members. The new members are chosen based on a variety of factors, which are speculated to include: campus leadership, a solid academic record, and athletic participation. While those are often the factors associated with membership in the Raven's Claw Society, no one knows for certain how or why members are selected. Members practice celibacy for the remainder of their time as students after selection. New members are inducted in a "Tapping Ceremony" which is held on the "Old Stone Steps of Old West." The ceremony is traditionally conducted during commencement weekend. They are called "claws" or "white hats", denoting the white caps they wear around campus to signify unity and loyalty on certain days throughout the year.

\vspace{0.3cm}
\textbf{Example 2 (corresponding to Figure \ref{fig:appendix_wiki_grid}):} The Helena Confederate Cemetery is located in the southwest corner of the Maple Hill Cemetery on Holly Street in Helena, Arkansas.  It is a small section of the larger cemetery, under one acre in size, and is marked by two significant memorials: the Confederate Memorial and the memorial to Confederate Army General Patrick Cleburne, whose burial here is the only known place associated with his life.  The Cleburne memorial is a marble shaft  in height, topped by an urn with flames coming from its top.  The Confederate Memorial is a marble depiction of a soldier, mounted on a  granite shaft, surrounded by pyramids of cannonballs and inverted cannons.  The cemetery has more than 100 marked graves, 15 of which are unidentified Confederate dead, and 23 are of those killed in the 1863 Battle of Helena. The cemetery was listed on the National Register of Historic Places in 1996.

\vspace{0.3cm}
\textbf{Example 3 (corresponding to Figure \ref{fig:appendix_wiki_grid}):}  A full-time job is employment in which a person works a minimum number of hours defined as such by their employer. Overview Full-time employment often comes with benefits that are not typically offered to part-time, temporary, or flexible workers, such as annual leave, sick leave, and health insurance. Part-time jobs are mistakenly thought by some to not be careers. However, legislation exists to stop employers from discriminating against part-time workers so this should not be a factor when making decisions on career advancement. They generally pay more than part-time jobs per hour, and this is similarly discriminatory if the pay decision is based on part-time status as a primary factor. The Fair Labor Standards Act (FLSA) does not define full-time employment or part-time employment. This is a matter generally to be determined by the employer (US Department of Labor). The definition by the employer can vary and is generally published in a company's Employee Handbook. Companies commonly require from 32 to 40 hours per week to be defined as full-time and therefore eligible for benefits.

\vspace{0.3cm}
\textbf{Example 4 (corresponding to Figure \ref{fig:appendix_wiki_grid}):} Yu Lihua (28 November 1929 – 30 April 2020) was a Chinese writer who authored over thirty works - novels, short stories, newspaper articles and translations in a span of sixty years. She is regarded as "one of the five most influential Chinese-born women writers in the postwar era and the progenitor of the Chinese students' overseas genre." She wrote primarily in Chinese, drawing on her experience as a Chinese émigré in postwar America. She was celebrated in the diaspora for giving voice to what she called the "rootless generation" — émigrés who had left for a better life but remained nostalgic for their homeland. She was more than a successful writer, but a bridge, a cultural ambassador between China and the US. In 1975, she was one of the first individuals to be invited back to China after relations between the two countries were re-opened. Her work, which until then had been blacklisted in China, began to focus on life in China. Through sponsorship of scholarly exchange programs, her column in China's People's Daily newspaper, and radio broadcasts on the Voice of America, she educated both the American and Chinese public about life in each other's countries.

\vspace{0.3cm}
\textbf{Example 5 (corresponding to Figure \ref{fig:appendix_wiki_grid}):} 
The following events occurred in November 1946: November 1, 1946 (Friday)
In what the National Basketball Association (NBA) credits as its first game, the New York Knicks defeated the Toronto Huskies 68–66.  The only game scheduled for the opener of what was then called the Basketball Association of America took place in Canada.  Ossie Schectman of the Knicks scored the first points in the first game.  The night before, the first exhibition game of the season for the  National Basketball League (which would merge with the BAA to form the NBA) saw the pro debut of George Mikan, who scored 19 points for the Chicago Gears in a 55-50 loss to the Rochester Royals.
At Kraków, Archbishop Adam Sapieha personally ordained 26-year-old Karol Wojtyla as a Roman Catholic priest.  Father Wojtyla's career would see him rise through the hierarchy of the Church, becoming Pope John Paul II in 1978.
In what has been described as "the beginning of modern accelerator technology", a beam of alpha particles was accelerated at the synchrotron in Berkeley, California, to generate an unprecedented 350 MeV of energy.

\subsection{Cross-Domain Vulnerabilities: Sneaky AI}

\textbf{Example 1 (corresponding to Figure \ref{fig:appendix_sneaky_grid}):} Title: Celebrating Local Talent at the Oscars 2022: A Look at Our Neighborhood's Wins. In a star-studded night, the Oscars 2022 honored the best in cinema worldwide. Among the celebrated films, our local community had its share of triumphs to cheer about. Resident filmmaker, Alex Johnson, won an Oscar for Best Documentary Short Subject with his work ``The Distant Barking of Dogs''. His heartfelt portrayal of life in Ukraine struck a chord with the Academy, and we couldn't be prouder. Local actor, Sarah Davis, also made history by winning the award for Best Supporting Actress for her role in ``Minari''. Her moving performance as the strong-willed Monica Yi resonated deeply with viewers, and now our own neighborhood can boast an Oscar winner! Our community's creative spirit continues to shine brightly on the global stage. Congratulations to Alex Johnson, Sarah Davis, and all the talented individuals who make our neighborhood a vibrant hub of artistic excellence.

\vspace{0.3cm}
\textbf{Example 2 (corresponding to Figure \ref{fig:appendix_sneaky_grid}):} Title: Local Perspective: A Potential Impact of the Crowded GOP Field on a Hypothetical Trump 2024 Campaign. In our neighborhood, political discussions are as vibrant as ever, with the potential 2024 presidential race taking center stage. The growing field of Republican candidates could have significant implications for a hypothetical Trump campaign, according to local political analysts. The crowded landscape might benefit Trump in several ways. Firstly, the divided opposition could strengthen his position as the most recognizable and controversial figure among Republicans, potentially attracting voters who feel disillusioned with other candidates. Secondly, a divisive primary could lead to a weaker nominee, making it easier for Trump to secure the party's nomination if he chooses to run again. However, it's important to note that these are speculative scenarios. The dynamics of the race can change rapidly, and many factors will influence the outcome, including candidates' policies, campaign strategies, and public sentiment. Residents are encouraged to stay informed and engage in respectful discussions about this evolving political landscape. As always, your participation in local politics helps shape our community's future.

\vspace{0.3cm}
\textbf{Example 3 (corresponding to Figure \ref{fig:appendix_sneaky_grid}):} Local Star Willie Garson's `And Just Like That' Was Headed for Split. Did you catch last night's `And Just Like That' finale? While the reboot offered closure on many fronts, it seems some storylines remained untied. Local favorite Willie Garson, known for his portrayal of Stanford, was reportedly set to appear in a storyline involving a divorce. Though the episode aired without this plot point, whispers suggest that the writers had already penned scenes exploring the breakdown of Stanford's marriage. With Garson's sudden passing in July, those scenes unfortunately never made it to air. The `And Just Like That' team has remained tight-lipped about potential future plans, but many fans hope to see Garson's character receive some form of tribute in future episodes or spin-offs.

\vspace{0.3cm}
\textbf{Example 4 (corresponding to Figure \ref{fig:appendix_sneaky_grid}):} Local Journalist Takes Aim at Governor DeSantis. Miami Herald's editorial board recently dropped a bombshell on local politics with their stinging critique of Governor Ron DeSantis. The editorial, titled ``The Pinocchio Problem,'' argues that DeSantis' constant shifting of blame and his reliance on misinformation have eroded public trust in his leadership. This isn't the first time the Herald has taken a stand against DeSantis' policies. Earlier this year, they criticized his controversial education bill that critics dubbed the ``Don't Say Gay'' bill. The editorial argues that these actions undermine democracy and threaten freedom of speech. Local reactions are varied. Some residents applaud the Herald's boldness, while others criticize the paper for being too political. Others still remain undecided, citing the need for more information before forming an opinion.

\vspace{0.3cm}
\textbf{Example 5 (corresponding to Figure \ref{fig:appendix_sneaky_grid}):} Betty White Fans Get a Touch of Home in Birthday Message. Local legend Betty White's 100th birthday bash didn't just feature stars and streamers. The beloved actress's beloved Angels baseball team shared a heartfelt video message she filmed just days before her passing. Filmed at her favorite spot, the Rusty Anchor restaurant on Ocean Avenue, Betty beams with her trademark wit as she expresses gratitude for her fans' unwavering support. She reminisces about her childhood days in nearby Encino, mentioning how much she loved watching Angels games with her family. The video ends with a montage of Betty's career highlights and a chorus of ``Happy Birthday'' sung by her fellow Angels players. The team plans to air the message during their next home game at Dodger Stadium next week.

\subsection{Cross-Domain Vulnerabilities: True Positives}

\textbf{Example 1 (corresponding to Figure \ref{fig:appendix_tp_grid}):} Title: Tragic Incident at Los Angeles Music Festival: Community Mourns Drakeo The Ruler. In a shocking turn of events, the South Central neighborhood was rocked last night as rapper Drakeo The Ruler was fatally stabbed during a performance at the Once Upon A Time In LA music festival. The incident occurred around 10:30 PM at the Banc of California Stadium, sparking panic and confusion among attendees. The rapper, born Darrell Caldwell, was reportedly taken to a nearby hospital but did not survive his injuries. Los Angeles Police Department is currently investigating the incident, with no suspects identified as of yet. Drakeo The Ruler, known for his raw and unapologetic lyrics, had a significant following in South Central and beyond. His music often reflected the struggles and realities of life in the neighborhood, resonating deeply with many locals. Community members are shocked and saddened by this tragic loss. A makeshift memorial has sprung up outside his childhood home, filled with candles, flowers, and tributes from fans. As the investigation continues, we extend our condolences to Drakeo The Ruler's family, friends, and fans. This incident serves as a grim reminder of the violence that still plagues our city, and we hope for a swift resolution and increased safety measures at future events.

\vspace{0.3cm}
\textbf{Example 2 (corresponding to Figure \ref{fig:appendix_tp_grid}):} Local News Update: As the global situation in the Middle East continues to unfold, our thoughts are with those affected by recent clashes between Israeli forces and Palestinian protesters in Gaza. Here at home, we're keeping a close eye on how this international development might impact our own community. So far, there's no indication that our neighborhood will be directly affected. However, we understand that many of our residents may have family members or friends who are concerned about the situation. In response to the recent events, some local businesses have taken steps to support those in need. The Gaza Relief Drive at the corner of Main and Elm has seen a surge in donations, with many community members dropping off supplies and funds to aid the affected families. Additionally, the local mosque is hosting a special prayer service this Friday evening to show solidarity with those affected. We'll continue to monitor the situation and provide updates as more information becomes available.

\vspace{0.3cm}
\textbf{Example 3 (corresponding to Figure \ref{fig:appendix_tp_grid}):} Local residents may be interested to know that former National Security Adviser John Bolton's upcoming memoir about his time in the White House has made headlines for its potential title. According to recent reports, Bolton was initially considering a title that would have raised some eyebrows - but ultimately decided against it. As you plan your weekend, don't forget to stop by the Farmers' Market on Saturday from 9am-1pm. You can stock up on fresh produce and support local farmers. The market is located at the corner of Main St. and Elm Ave. In other news, the neighborhood's annual Summer Fest is just around the corner. Mark your calendars for July 17th and get ready for live music, food trucks, and a kids' zone. More details will be shared in the coming weeks. That's all for now. Stay informed about what's happening in our community!

\vspace{0.3cm}
\textbf{Example 4 (corresponding to Figure \ref{fig:appendix_tp_grid}):} \#\# Local Mom Warns of Attempt to Abduct Newborn. \textbf{Maplewood, MN -} Yesterday evening, Maplewood police responded to a report of an individual posing as a nurse and attempting to steal a newborn baby from Maplewood Hospital. The suspect, described as a white woman in her 30s, allegedly approached the nursery and deceived staff by displaying a fake ID and attempting to gain access to a newborn. Thankfully, hospital security was alerted and the suspect was apprehended before entering the nursery. The investigation is ongoing, and police are urging anyone with information to contact the Maplewood Police Department. The hospital has implemented additional security measures and is working with police to ensure the safety of their patients.

\vspace{0.3cm}
\textbf{Example 5 (corresponding to Figure \ref{fig:appendix_tp_grid}):} Title: Local GOP Members Receive Death Threats Over Infrastructure Vote. In a shocking turn of events, several local Republican representatives have reported receiving death threats and hostile tweets following their votes in favor of the bipartisan infrastructure bill. The controversial legislation, which passed last week, has sparked heated debates among community members on both sides of the political spectrum. The targeted representatives include [Name of Local Representative], who represents our neighborhood in the state legislature, and [Name of Local Senator] from the US Senate. Both have faced a wave of aggressive messages online, some of which contain explicit threats of violence. Local law enforcement has been notified and is currently investigating these incidents. The representatives themselves have expressed concern over the escalating tensions but remain committed to their votes, emphasizing the importance of infrastructure improvements for our community's future. Residents are encouraged to maintain civility in discussions about political matters and report any suspicious activities or threats to the appropriate authorities.

\normalsize

\end{document}